\documentclass{article}
\PassOptionsToPackage{table}{xcolor}
\usepackage{iclr2027_conference,times}
\usepackage{iclr2027_arxiv}

\usepackage[T1]{fontenc}
\usepackage[utf8]{inputenc}

\usepackage{amsmath,amsfonts,bm}

\def\eqref#1{equation~\ref{#1}}

\def\1{\bm{1}}

\DeclareMathAlphabet{\mathsfit}{\encodingdefault}{\sfdefault}{m}{sl}
\SetMathAlphabet{\mathsfit}{bold}{\encodingdefault}{\sfdefault}{bx}{n}

\usepackage{tocloft}
\usepackage[hypertexnames=false]{hyperref}
\usepackage{url}
\usepackage{booktabs}
\usepackage{multirow}
\usepackage{xspace}
\usepackage{enumitem}
\usepackage{graphicx}
\usepackage{caption}
\usepackage{float}
\usepackage{cleveref}
\usepackage{etoc}

\AddToHook{cmd/appendix/before}{%
  \crefalias{section}{appendix}%
  \crefalias{subsection}{subappendix}%
  \crefalias{subsubsection}{subsubappendix}%
}
\usepackage{xcolor}
\usepackage[table]{xcolor}
\usepackage{pifont}

\newcommand{\ymark}{\ding{51}}
\newcommand{\xmark}{\ding{55}}

\definecolor{techblue}{RGB}{39,117,195}
\definecolor{techgray}{RGB}{240, 240, 240}
\definecolor{techdark}{RGB}{30, 30, 30}

\newcommand{\framework}{BLIP-Live\xspace}
\newcommand{\frameworkFull}{BLIP-Live\xspace}
\newcommand{\dataset}{BLIP-Live-780K\xspace}
\newcommand{\gemmaModel}{BLIP-Live-Gemma\xspace}
\newcommand{\qwenModel}{BLIP-Live-Qwen\xspace}

\definecolor{krgreen}{RGB}{27, 122, 78}

\makeatletter
\renewcommand\paragraph{\@startsection{paragraph}{4}{\z@}%
  {0.2ex plus 0.5ex minus .2ex}{-1em}{\normalsize\bf}}
\makeatother

\title{BLIP-Live: Balancing Timeliness \& Firing Precision in Proactive Streaming Video Agents}

\author{%
    Kanchana Ranasinghe \qquad Honglu Zhou \qquad Dominick Reilly \qquad Le Xue
    \\
    \qquad Shusheng Yang \qquad Juan Carlos Niebles \qquad Silvio Savarese \qquad Ran Xu \\
    Salesforce AI Research
}

\seticlrarxivauthors{%
  Kanchana Ranasinghe, Honglu Zhou, Dominick Reilly, Le Xue, Shusheng Yang, Juan Carlos Niebles, Silvio Savarese, Ran Xu%
}
\seticlrarxivaffiliation{Salesforce AI Research}
\seticlrarxivwebsite{https://bliplive.salesforceresearch.ai}

\hypersetup{colorlinks=true,linkcolor=techblue,citecolor=techblue,urlcolor=techblue,
  pdftitle={Balancing Timeliness and Firing Precision in Proactive Streaming Video Agents},
  pdfauthor={Kanchana Ranasinghe, Honglu Zhou, Dominick Reilly, Le Xue, Shusheng Yang, Juan Carlos Niebles, Silvio Savarese, Ran Xu}}

\iclrarxivcopy

\begin{document}
\etocdepthtag.toc{main}
\maketitle

\begin{abstract}
Streaming video understanding is becoming increasingly important for real-world interactive systems.
Such systems must be \textit{proactive} for natural and useful human interaction: 
they should respond as soon as sufficient evidence becomes available while 
avoiding premature, redundant, or incorrect responses.
We present \frameworkFull, a framework unifying data, metrics, harness, and training around a shared formulation of proactivity as \textit{effective causal response decisions} balancing the two competing objectives of responding vs silence.  
First, we construct \dataset dataset containing human annotations of proactive response content and timing, which defines this desired behavior. 
We then propose PAUC-F, a metric to effectively measure our desired proactive behavior. 
PAUC-F combines existing response timeliness metrics with firing precision to reward early useful responses \textit{without} rewarding excessive firing. 
With our desired behavior defined and measurable, we finally investigate model-agnostic inference harnesses that draw out proactive streaming capabilities of offline vision language models (VLMs), together with training pipelines that learn to further improve their proactive responsiveness.
Across both settings, we find that combinations of sliding visual windows, online interaction memory, and structured response outputs consistently yield strongest proactive performance.
Our experiments result in two models, with our \qwenModel model achieving state-of-the-art performance across four proactive streaming benchmarks, with gains up to 44\% over baselines and outperforming multiple closed API models. 
Our code and data will be released publicly.
\end{abstract}

\section{Introduction}

Proactive streaming video agents must learn not only what to say, but also \textit{when to say it}, and \textit{when to remain silent} \citep{wang2026mmduet2,wang2025proactivevqa}.
This differs from offline video question-answering, where queries are immediately followed by responses. 
Streaming agents observe videos incrementally (causal), and at each frame decide whether to respond or remain silent for a query provided at an earlier time \citep{chen2024videollmonline,zhang2025eyeswideopenego}.
This causal firing decision presents a key challenge: a correct response is unhelpful if it arrives too late or is repeated too often \citep{wang2026mmduet2}. 
Effective proactivity requires balancing response timeliness with firing precision.

\begin{figure}[t]
  \centering
  \includegraphics[width=\linewidth]{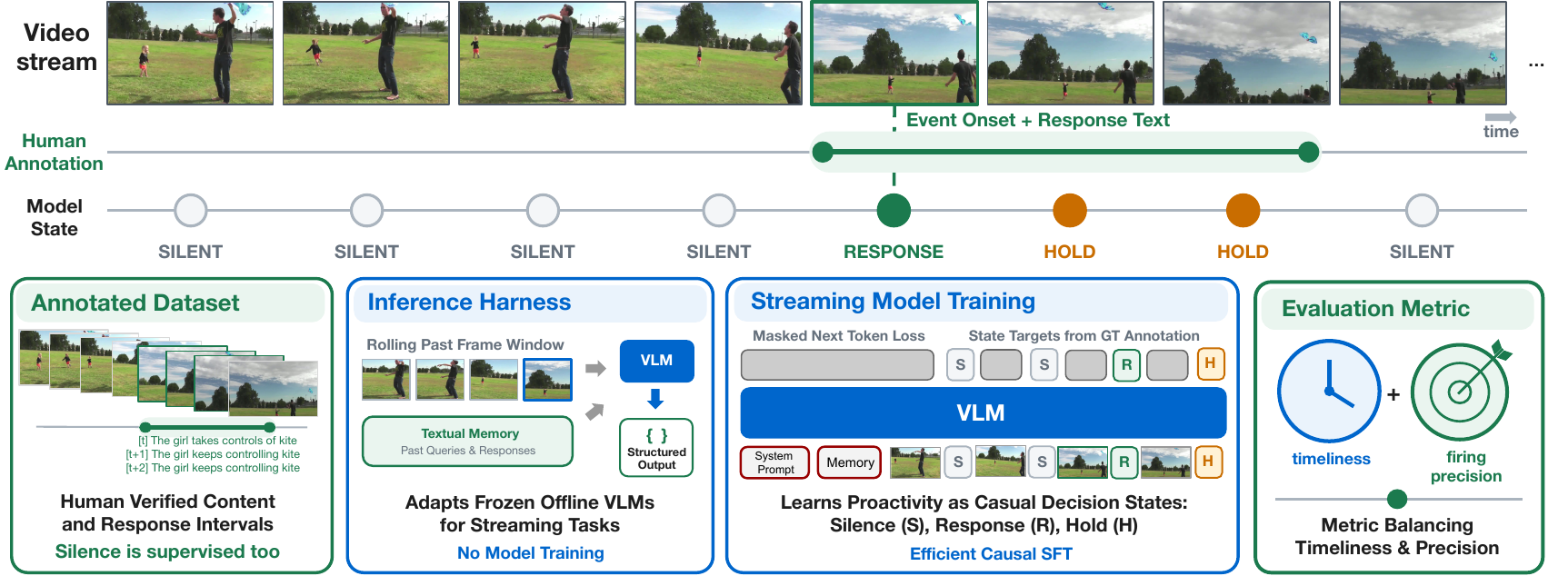}
  \vspace{-1.5em}
  \caption{
  \textbf{Overview of proposed \framework framework.} 
  We unify data, metrics, harness, and training around a shared formulation of proactivity as \textit{effective causal response decisions} balancing the two competing objectives of responding vs silence in the proactive streaming task.  
  }
  \label{fig:method_overview}
  \vspace{-1.0em}
\end{figure}

In order to achieve this desired behavior in streaming video agents, we propose aligning the end-to-end pipeline of data, metrics, harness, and training under a shared formulation.
While existing works seperately address subsets of these, 
we present \frameworkFull, \underline{B}ootstrapping \underline{L}ive \underline{I}nteractive \underline{P}erception, a single framework aligning data, metrics, harness, and training around a shared formulation of proactive responding as effective causal response decisions (illustrated in \Cref{fig:method_overview}).

We first construct \dataset, a large-scale dataset of proactive responses built through automated proactive question-answer generation followed by comprehensive human verification. 
In contrast to existing proactive streaming datasets (\Cref{tab:dataset_comparison}), \dataset provides human verified response content and intervals, providing explicit human supervision on when to remain silent. 
Next, we propose PAUC-F, an evaluation metric that captures both timeliness and firing precision in proactive streaming model responses. 
Prior metrics such as PAUC \citep{wang2025proactivevqa} and ESTP-F1 \citep{zhang2025eyeswideopenego} separately measure individual aspects of the proactive streaming task, leaving room for models to game these metrics with undesirable traits (see \Cref{fig:pauc_vs_paucf} and \Cref{tab:zss5_zss7_selection}).
Our proposed PAUC-F, retains PAUC's quality-weighted timeliness and balances it with response precision.

With desired behavior defined and measurable, next we investigate harness designs and training pipelines that improve proactive behavior in VLMs. 
Existing works explore proactivity with learned activation models \citep{wang2025streambridge}, hand-crafted harness designs \citep{yao2026harnessing}, learned response decision tokens \citep{chen2024videollmonline,li2026videochat3}, and response policy optimization \citep{wang2024mmduet,wang2026mmduet2}.
We introduce a model-agnostic harness tuning strategy to elicit proactivity in VLMs. 
Our selected harness combines a sliding visual window, interaction memory, and structured decision states.
The same interface then supports efficient teacher-forced training with our datasets, further improving proactive responsiveness in VLMs, and resulting in two models, \gemmaModel and \qwenModel. 
Our \qwenModel model achieves state-of-the-art performance across four proactive streaming benchmarks, outperforming multiple closed source models as well.

We summarize our key contributions as follows:
\begin{enumerate}[leftmargin=1.8em,topsep=-0.5ex,itemsep=0.0ex,partopsep=0ex]
  \item \textbf{\dataset Dataset:} A high-quality proactive streaming dataset for training built through automated generation of proactive question–response pairs, followed by comprehensive human review and filtering to ensure accurate response content and timing annotations.
  \item \textbf{PAUC-F Metric:} Evaluation metric designed to measure proactive understanding in VLMs that captures both response timeliness \textit{and} firing precision.
  \item \textbf{Inference Harness:} Model-agnostic harness tuning and calibration that adapts offline VLMs for proactive responding under streaming style inference.
  \item \textbf{Training Framework:} Efficient training pipeline resulting in \gemmaModel and \qwenModel models that yield relative improvements up to 44\% over prior baselines across benchmarks.
\end{enumerate}

\section{Related Work}

\paragraph{Proactive Video Understanding.}
Proactive responsiveness is tackled under two key directions: inference harness and training pipeline.  
Harness design involves response gates, event detection, and long-context memory management \citep{wang2025streambridge,qian2025dispider,ding2025streammind,azad2026streamready,zhang2026progressive,tian2026roma,yao2025timechatonline,zheng2026emgarde,yao2026harnessing}, but its role is mainly streaming setting adaptation, overlooking direct optimization for proactivity.
Proactive behavior is handled mostly in model through response decision states \citep{chen2024videollmonline,xia2026streamingvideo,li2026videochat3}, implicit signals in interleaved video-text data \citep{wang2024mmduet,yang2025livestar}, and reinforcement learning \citep{wang2026mmduet2}.
In contrast, our proposed harness is model agnostic and tuned specifically for proactivity, while the same interface extends to supervision when training data is available and transfers across model families rather than committing to fixed architectures.

\paragraph{Proactive Streaming Datasets.}
Prior works convert offline annotations from existing datasets to streaming style, often with LLM assistance. 
\citet{chen2024videollmonline} reformats COIN and Ego4D narrations directly into streaming dialogues, 
while \citet{wang2024mmduet,wang2025streambridge} use dense captioning task data to determine response timing using annotated temporal segments. 
\citet{wang2026mmduet2,yao2025timechatonline} generate dialogue from scene captions with text-only LLMs. 
\citet{xia2026streamingvideo,li2026videochat3} derive labels from a VLM at scale. 
However, existing datasets suffer from poor response timing \citep{wang2026mmduet2}, affecting quality of proactive responsiveness. 
In contrast, \dataset ensures accurate response content and timing through comprehensive human verification (see \Cref{tab:dataset_comparison}).

\paragraph{Evaluation Metrics for Proactive Understanding.}
Early benchmarks evaluate proactivity through targeted tasks. 
OVO-Bench \citep{li2025ovobench} introduces  CRR (Clue Reveal Responding) task where the same query requires different responses at different timesteps. 
StreamingBench proposes Proactive Output (PO) task that evaluates the first response occurring within $\pm2$ seconds of an annotated timestamp \citep{lin2024streamingbench}. 
In contrast, later ESTP-F1 \citep{zhang2025eyeswideopenego} and PAUC \citep{wang2025proactivevqa} metrics evaluate continuous interaction under true streaming settings, but both overlook firing-precision to certain extents (\Cref{fig:pauc_vs_paucf}), motivating our proposed PAUC-F metric.

\paragraph{Automated Harness Design.} Evolving inference harnesses is explored for general agents and offline video understanding \citep{zhang2026darwinx,xu2026videoharnessrsi,zhang2026harnesscompass}, but ours is a first to demonstrate successful integration to proactive or streaming video tasks.

\section{The \dataset Dataset}
\label{sec:dataset}

We construct a dataset for learning proactive response behavior in streaming video tasks, containing 784,944 proactive responses across 177,789 videos.
Each video includes a user query followed by a sequence of temporally grounded natural-language responses. Our dataset uses the same source videos as LLaVA-Video-178K \citep{zhang2024llava}. We employ a three-stage annotation pipeline:
\begin{itemize}[label={},nosep,leftmargin=1.8em,topsep=0pt,itemsep=1pt,parsep=0pt]
    \item \textbf{Stage 1: Automated Generation.}
    Creates candidate proactive query and response sequences.
    \item \textbf{Stage 2: Human Verification and Correction.}
    Human annotators inspect generated samples and correct errors related to both response timing and content. 
    \item \textbf{Stage 3: Expert Quality Review.}
    Independent PhD-level reviewers evaluate corrected annotations in batches. 
    Batches below acceptance threshold are returned to Stage~2 for further revision.
\end{itemize}

Note that each query can require a single or multiple responses across the video, with each response corresponding to a single frame or frame interval. 
Our annotations capture both response and silent timings to encourage firing precision. 
\Cref{tab:dataset_comparison} shows how compared to earlier work, our dataset is larger in scale and provides human-verified annotations for both response content and intervals.

\begin{table}[t]
\caption{
\textbf{Proposed Streaming Proactive Dataset:}
We construct a training dataset for proactive streaming task with human-annotated content and intervals.}
\label{tab:dataset_comparison}
\vspace{-0.5em}
\centering
\scriptsize
\setlength{\tabcolsep}{8pt}
\begin{tabular}{@{}lrrrcc@{}}
\toprule
\multirow{2}{*}{Dataset} & \multicolumn{1}{c}{\multirow{2}{*}{Videos}} & \multicolumn{1}{c}{\multirow{2}{*}{Questions}} & \multicolumn{1}{c}{\multirow{2}{*}{Responses}} & \multicolumn{2}{c}{Human-verified} \\
\cmidrule(l){5-6}
& \multicolumn{1}{c}{} & \multicolumn{1}{c}{} & \multicolumn{1}{c}{} & Content & Intervals \\
\midrule
MMDuet2~\citep{wang2026mmduet2} & 52,771 & 105,800 & 350,800 & No & No \\
VideoChat3-OL~\citep{li2026videochat3} & 191,000 & 617,183 & 617,183 & No & No \\
ESTP-IT~\citep{zhang2025eyeswideopenego} & 1,166 & 80,000 & 80,000 & No & No \\
PARSE-Ego4D~\citep{abreu2025parseego4d} & 5,515 & 18,360 & 18,360 & Yes & No \\ \midrule
\rowcolor{techgray}
\textbf{\dataset (Ours)} & \textbf{177,789} & \textbf{177,789} & \textbf{784,944} & \textbf{Yes} & \textbf{Yes} \\
\bottomrule
\end{tabular}
\vspace{-1.5em}
\end{table}

\paragraph{Stage 1: Automated Proactive Query-Response Generation.}

We generate proactive interactions for a given video using a VLM in two successive stages.
First, for query generation, we provide the VLM with a single video and obtain a single video-specific user query relevant to measuring proactive responsiveness. 
Next, for response generation, we provide the VLM both the video and query to obtain an ordered sequence of response or silence states. This is generated at 1 FPS, i.e. one state per second.  
Response states also include the response content. 
The generator receives the full video but is prompted to use only evidence available up to each response timestamp; humans then verify and correct timing and content. 

\paragraph{Stage 2: Human Verification and Correction.}
Human annotators first filter videos based on query validity, and then evaluate each response or silent state. Incorrect states are fixed, including correction of response content when necessary.  
Query validity is defined as requiring only visually verifiable conditions and unambiguous answers. 
Response and silent states are evaluated based on alignment with evidence present in video until the relevant timestep, and corrected accordingly. 
Note that response states capture not only instantaneous events but also continuous events which involve consecutive response states over multiple timesteps.
Of the responses generated, human review modified ~11\% and removed ~40\%, while adding ~28\% new responses. We provide full editing and filtering statistics in \Cref{tab:annotation_edits,tab:annotation_outcomes}.

\begin{table}[t]
\centering
\begin{minipage}[t]{0.44\linewidth}
\vspace{0pt}
\caption{\textbf{Dataset Filtering.} Generated and human edited (final) record counts; net reductions include deduplication.}
\label{tab:annotation_edits}
\vspace{-0.5em}
\centering
\small
\setlength{\tabcolsep}{3pt}
\def\arraystretch{1.2}  
\scalebox{0.85}{
\begin{tabular}{lrrrr}
\toprule
Records & Generated & Final & \shortstack{Net\\reduction} & \shortstack{Reduction\\(\%)} \\
\midrule
Queries & 186,751 & 177,789 & 8,962 & 4.80 \\
Responses & 893,562 & 784,944 & 108,618 & 12.16 \\
\bottomrule
\end{tabular}}
\end{minipage}\hfill
\begin{minipage}[t]{0.54\linewidth}
\vspace{0pt}
\caption{\textbf{Annotation outcomes.} Percentages relative to the 893,562 response records from automated generation stage.}
\label{tab:annotation_outcomes}
\vspace{-0.5em}
\centering
\small
\setlength{\tabcolsep}{1.2em}
\def\arraystretch{1.1}  
\scalebox{0.85}{
\begin{tabular}{lrr}
\toprule
Outcome & Count & \% of generated \\
\midrule
Unchanged & 427,150 & 47.80 \\
Modified & 104,468 & 11.69 \\
Deleted & 361,944 & 40.51 \\
Added & 253,326 & 28.35 \\
\bottomrule
\end{tabular}}
\end{minipage}
\end{table}

\paragraph{Stage 3: Quality Control.}
All annotators are recruited and trained by a professional video data annotation team. 
We provide detailed instructions and visual examples describing our annotation requirement. 
The annotated samples undergo quality control at batch level, using PhD-level expert reviewers (who designed the annotation process). 
After each batch, annotation proceeds to next only if quality exceeds our required threshold. 
Failing batches are returned for further review and correction. 
While initial batches required one repeat cycle, later batches passed required quality directly, ascertaining that annotation process was well aligned with our specified requirement.

\section{PAUC-F: Proactive Streaming Evaluation Metric}
\label{sec:paucf}

We propose PAUC-F that jointly captures answer quality, timeliness, and firing precision. 
In \Cref{tab:zss5_zss7_selection}, humans prefer a harness optimized with this PAUC-F metric over 90\% of the time.

\subsection{Limitations of Prior Metrics}

\begin{figure}[t]
  \begin{minipage}[t]{0.65\linewidth}
  \centering
  \includegraphics[width=\linewidth]{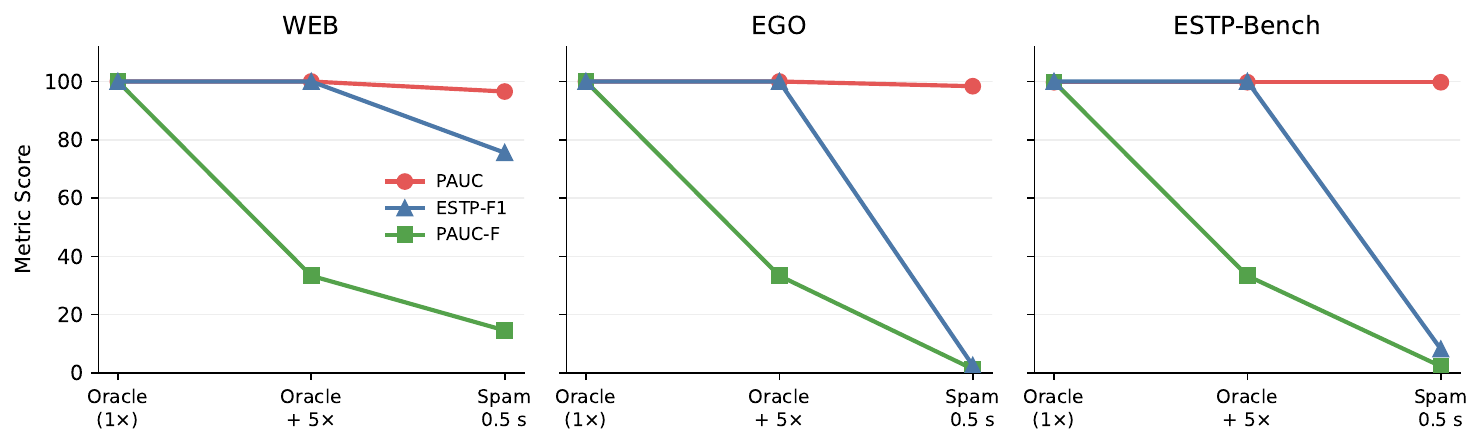}
  \vspace{-2.0em}
  \caption{
  \textbf{PAUC-F vs Prior Metrics:}
  We simulate excessive firing behaviour (undesired) across WEB \& EGO splits of ProactiveVideoQA \citep{wang2025proactivevqa} and Explicit Proactive subset of ESTP-Bench \citep{zhang2025eyeswideopenego} benchmarks, reporting our proposed PAUC-F metric alongside PAUC \& ESTP-F1 metrics.
  Along the x-axes, we repeat ground-truth responses (Oracle) 5 additional times within golden windows (Oracle + 5x) and every 0.5s throughout the video (Spam 0.5).
  PAUC-F correctly identified these undesired response patterns while PAUC and ESTP-F1 incorrectly rewards all or some of these undesired test cases.}
  \label{fig:pauc_vs_paucf}
  \end{minipage}
  \hspace{0.01\linewidth}
  \begin{minipage}[t]{0.28\linewidth}
  \centering
  \vspace{-7em}
  \captionof{table}{\textbf{PAUC-F vs PAUC for model selection.}
  We build two harness variants optimizing PAUC (v1) vs PAUC-F (v2). 
  PAUC-F optimized v2 shows better firing precision, closer firing counts as ground-truth (Fires:Gold), less consecutive repeated responses, and higher human preference.  
  }
  \label{tab:zss5_zss7_selection}
  \vspace{-0.5em}
  \centering
  \small
  \setlength{\tabcolsep}{5pt}
  \def\arraystretch{1.2}
  \resizebox{\linewidth}{!}{%
  \begin{tabular}{@{}lcc|cc@{}}
  \toprule
  & \multicolumn{2}{c|}{WEB} & \multicolumn{2}{c}{EGO} \\
  \cmidrule(lr){2-3}\cmidrule(l){4-5}
  Metric & v1 & v2 & v1 & v2 \\
  \midrule
  PAUC ($\omega=0.5$) $\uparrow$   & \underline{64.82} & 57.30 & \underline{66.57} & 37.57 \\
  PAUC-F ($\omega=0.5$) $\uparrow$ & 31.92 & \underline{43.39} & 3.58 & \underline{35.94} \\
  Firing Precision $\uparrow$      & 21.17 & \underline{34.91} & 1.84 & \underline{34.45} \\
  Fires : Gold $\downarrow$        & 4.50$\times$ & \underline{2.29$\times$} & 48.18$\times$ & \underline{0.81$\times$} \\
  Repeat Responses $\downarrow$    & 14.30 & \underline{2.20} & 185.90 & \underline{1.32} \\
  Human Preference $\uparrow$      & 6\% & \underline{94\%} & 3\% & \underline{97\%} \\
  \bottomrule
  \end{tabular}%
  }
  \end{minipage}
\end{figure}

PAUC \citep{wang2025proactivevqa} and ESTP-F1 \citep{zhang2025eyeswideopenego} are two key metrics in recent literature relevant to proactivity in streaming video agents. 
PAUC measures quality weighted timeliness in responses. 
However, PAUC only measures recall within the gold-window. This results in responses outside of the gold-window window being discarded, and repeated responses within the window not affecting the score (see Figure~\ref{fig:pauc_vs_paucf}).
Similar observations are noted in MMDuet2 \citep{wang2026mmduet2}, where removing the repetition reward increases PAUC scores although duplicate replies increase by 4x. 
ESTP-F1 addresses out-of-window firing with the addition of a precision term, but ignores redundant in-window firing and requires task-dependent timing term for recall calculation.
These two key metrics provide complementary but incomplete views, which we demonstrate in \Cref{fig:pauc_vs_paucf}.

\subsection{PAUC-F Definition}
We propose PAUC-F, which maintains PAUC as its recall and introduces a precision term:
\begin{equation}
  R = \mathrm{PAUC}, \qquad 
  P = \frac{N_{\mathrm{value}}}{N_{\mathrm{fires}}}, \qquad
  \mathrm{PAUC\text{-}F} = \frac{2PR}{P+R},
\end{equation}
Here $N_{\mathrm{fires}}$ counts all emitted responses, including out-of-window fires. Each gold turn contributes at most one value-adding response to $N_{\mathrm{value}}$: its earliest in-window fire with judge score 1 or 2 (maximum $S=2$). Additional fires increase only the denominator. PAUC-F inherits PAUC's quality--timeliness weight $\omega$ and penalizes unnecessary firing through precision. Aggregation, zero-fire handling, and PAUC's $[0,1]$ range (with no-reply score 0.25) are detailed in \Cref{app:paucf}.

\subsection{Metric Properties}

We investigate properties of our proposed PAUC-F metric in \Cref{fig:pauc_vs_paucf} and \Cref{tab:zss5_zss7_selection}. 
First, we construct a toy experiment simulating undesired excessive firing behavior. 
We use ground-truth response content and intervals to mimic models exhibiting within golden window repetitions and excessive out-of-window firing. 
As observed in \Cref{fig:pauc_vs_paucf}, both PAUC and ESTP-F1 fail to capture at least one of these undesired repetitive response behaviors, while PAUC-F correctly penalizes both. 
This excessive firing simulation was motivated by findings in prior work where models optimized for PAUC alone exhibit such repetitive behavior in their responses \citep{wang2026mmduet2}. 

We also observe similar findings in our experiments, which we summarize in \Cref{tab:zss5_zss7_selection}. 
We optimize two inference harness variants (details in \Cref{sec:inference_harness}) to independently maximize PAUC (\texttt{v1}) and PAUC-F (\texttt{v2}) on a held-out validation set of our \dataset dataset. 
PAUC-F optimized \texttt{v2} leads to clearly more desired behavior on downstream tasks compared to the PAUC optimized \texttt{v1} agent. 
We also conduct a human preference study involving 10 participants rating 20 samples from WEB and EGO splits of \citep{wang2025proactivevqa}.
Each sample contains response predictions from \texttt{v1} and \texttt{v2} agents that are rated for better proactive response quality. 
Human raters show clear preference for the PAUC-F optimized \texttt{v2} agent (more details in \Cref{app:paucf}).

We further highlight how PAUC-F better captures the range of outcomes worse than silence (incorrect responses that reduce precision). \Cref{tab:sparse_policy} illustrates this, where additional out-of-window fires leave PAUC unchanged while reducing PAUC-F.

\begin{table}[t]
\centering
\caption{\textbf{Sparse-response example over 100 gold turns.} A single perfect response at a window's start raises mean PAUC to $(1+99\times0.25)/100=0.2575$. Additional out-of-window fires leave PAUC unchanged but reduce PAUC-F. Scores are on $[0,1]$.}
\label{tab:sparse_policy}
\vspace{-0.5em}
\small
\def\arraystretch{1.2}
\setlength{\tabcolsep}{1.1em}
\scalebox{0.85}{
\begin{tabular}{lrrr}
\toprule
Policy & PAUC & Precision & PAUC-F \\
\midrule
One perfect response; otherwise silent & 0.2575 & 1.0000 & 0.4095 \\
Same response + 1 out-of-window fire & 0.2575 & 0.5000 & 0.3399 \\
Same response + 5 out-of-window fires & 0.2575 & 0.1667 & 0.2024 \\
Same response + 10 out-of-window fires & 0.2575 & 0.0909 & 0.1344 \\
Same response + 100 out-of-window fires & 0.2575 & 0.0099 & 0.0191 \\
\bottomrule
\end{tabular}}
\end{table}

\section{Method}
\label{sec:method}

In this section, we introduce our inference harness and training pipeline for streaming video tasks.

\paragraph{Streaming Interaction Protocol.}
We first establish our definition of streaming inference, which is aligned with prior work \citep{chen2024videollmonline,wang2026mmduet2,li2026videochat3}. 
At time $t$, a model receives a window of past video frames and a textual state capturing prior queries and responses. 
In turn, the model must produce either silence or a response at the current timestamp. 
The model cannot access future frames, revise an earlier response, or assign a new answer to an older timestamp. 
This protocol aligns with the same prior work and allows models to understand streaming videos extending beyond model visual context and to proactively respond at any time after a user query.

\paragraph{Streaming Video Model.}
We also define a \textit{streaming video model} as one trained specifically to consume inputs and generate outputs in this format, as opposed to offline VLMs trained for offline video question-answering that can be adapted to this task through suitable harness design. 

\subsection{Inference Harness}
\label{sec:inference_harness}

Prior works often measure proactive behaviour of offline VLMs using non-streaming inference settings \citep{wang2025proactivevqa,chen2024videollmonline,zhang2025eyeswideopenego}, making direct comparison to streaming models difficult. 
For example, \citet{wang2025proactivevqa} evaluates offline VLMs on 5s frame chunks with responses of frame $t$ conditioned on frames up to $t+5$, deviating from the streaming task setting. 
More recent works integrate offline VLMs with an inference harness \citep{li2026videochat3,wang2025streambridge,yao2026harnessing} for evaluation under true streaming settings. 
While these works hand-craft harness designs, we focus on tuning a harness, motivated by recent work on automated harness evolution \citep{zhang2026darwinx}, to elicit stronger proactive responsiveness from an offline model. 

\begin{figure}[t]
\centering
\includegraphics[width=\linewidth]{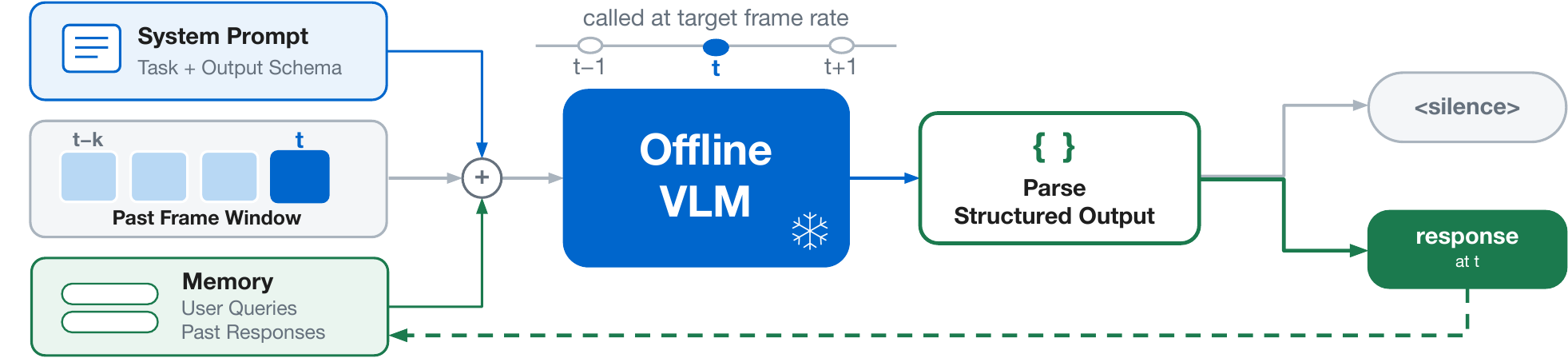}
\caption{
\textbf{Inference Harness.}
At each streaming timestep $t$, the harness constructs an input from a system prompt (task and output schema), a rolling past-frame window, and memory of user queries and prior responses. A frozen offline VLM is queried at the target frame rate; structured output is parsed into either silence or a response at $t$, and accepted responses are updated into memory.}
\label{fig:inference_harness}
\vspace{-1.0em}
\end{figure}

\paragraph{Design Overview.}
As illustrated in \Cref{fig:inference_harness}, our harness contains four key components: system prompt defining task and output schema, past frame window, memory holding prior interactions, and structured output parsing.
The underlying model is called at the desired frame rate, with dynamic inputs at each timestep conditioned on the past model responses.

\paragraph{Harness Tuning.} 
We next optimize the design of its four key components. 
Instead of hand-crafting these design choices, we utilize agent-assisted search to modify the harness, keeping underlying VLM model weights frozen (i.e. no model training). 
The agent first proposes hypotheses for modifying one of these four components, implements relevant code changes, and evaluates on \dataset validation set using our PAUC-F metric. 
These results guide new hypothesis creation over several rounds and the final design selection. 
Each round contains ten candidate variants and we promote variants that improve our objective (PAUC-F on validation set). 
Combination of complementary promoted changes is treated as a novel hypothesis for next round. 
We run 10 rounds of propose-evaluate-promote cycles with a total of 100 candidate evaluations each for two different models: \texttt{Gemma4-26B-A4B} \citep{gemmateam2026gemma4} and \texttt{Qwen3.5-35B-A3B} \citep{qwenteam2026qwen35}.
Both variants converge to a common design with minimal variations in system prompt and output schema. 

\paragraph{Harness Calibration.}
Given our fixed harness, we perform further model-specific calibration of system prompt and output schema with a more modest compute budget, using the same validation set and PAUC-F metric. 
We evaluate five candidates on a fixed 20 sample subset, with each candidate modifying only the system prompt and output schema (joint changes allowed). 
This corresponds to 5 rounds (one candidate per round) and a total budget of evaluating 100 video samples. 
We perform this limited calibration on all other offline models prior to evaluation.

\paragraph{Interactivity:}
The user query is included in our memory which can be updated midway during a video. 
This design choice allows user interaction with the framework mid-video: a human user can pose new queries as the framework continues processing an existing video, further aligning with true streaming setting.
We investigate and evaluate this behavior further in \Cref{tab:sb_po}. 

See \Cref{app:harness} for further details of our training-free harness design and its prompts. 

\subsection{Streaming Model Training}
\label{sec:streaming_training}

\begin{figure}[t]
\centering
\includegraphics[width=\linewidth]{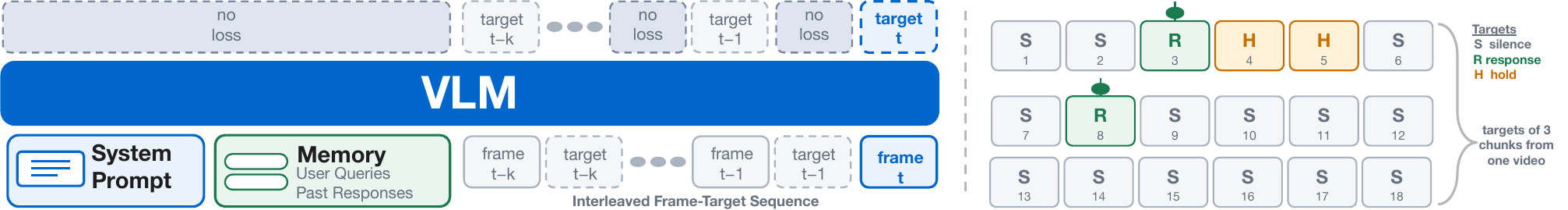}
\vspace{-1.5em}
\caption{
\textbf{Streaming Model Training.}
Each training sequence concatenates a system prompt, memory of user queries / past responses, and an interleaved frame-target stream.
Prompt, memory, and visual tokens receive no loss; the VLM is supervised on state targets (silence, response, and hold) and the accompanying response text. One video is split into multiple chunks so that long streams remain within the training context.}
\label{fig:streaming_training}
\vspace{-0.5em}
\end{figure}

We next investigate how supervised finetuning (SFT) can further improve proactivity. 
We use streaming data derived from \dataset and construct a training pipeline that retains the efficiency of standard teacher-forced SFT.
An overview of our training formulation is illustrated in \Cref{fig:streaming_training}.

\subsubsection{Streaming Episodes and Targets}
\label{sec:causal_episodes}

\paragraph{Causal episode construction.}
We construct dense sequences of frame-level decisions from our annotated videos.
First we build full video timelines followed by division into bounded chunks. 
A video timeline is a sequence of timestamped frame inputs and model turns, where model turns contain one of three state tokens,
\texttt{</Silence>}, \texttt{</Response>}, or \texttt{</Hold>}. 
The response token is followed by the annotated natural-language response. 
These timelines are split into chunks that contain the memory, absolute timestamps, and its corresponding source-frame range.
Within chunks, all response turns are fully supervised, while long runs of non-response states are subsampled to avoid silence dominating the token budget.
A chunk corresponds to $L=32$ frames, which is the past frame window that is provided to our model. 

\paragraph{State targets.}
Human annotations define three states: irrelevant frames are \texttt{Silence}; the first frame with sufficient evidence is \texttt{Response}; subsequent frames of the same ongoing event are \texttt{Hold}, suppressing repetition. Only $18\%$ of responses are followed by Hold. Post-response silence-only chunks teach sustained silence. Finetuned models predict all three states at inference; Hold and Silence emit no response and leave memory unchanged. The frozen-model harness uses binary silence/response outputs (\Cref{tab:final_harness_config}).
See \Cref{app:training} for further details on state targets and our training data distribution.

\subsubsection{Streaming Supervision}
\label{sec:streaming_supervision}

We perform teacher-forcing SFT by using individual chunks as samples.
Each chunk contains a system prompt, memory, and an interleaved sequence of frame-target pairs. 
The targets are the states \texttt{</Silence>}, \texttt{</Response>}, or \texttt{</Hold>}.  

\paragraph{Memory Module.}
Each training chunk includes the user query and last $m=1$ annotated response; inference uses the model's last generated response. Memory retains responses outside the visual window to discourage repetition, and new queries can update it mid-video. Both finetuned models use $L=32$ frames and $m=1$ during training and inference. The frozen-model harness uses three frames and $m=1$; both settings run at 1\,Hz.

\paragraph{Loss Masking.}
While our training employs a standard next token prediction loss, within each chunk we mask the cross-entropy loss term at token level. 
This is motivated by the large portion of silence timestamps inherent to the task and similar strategies in prior work \citep{li2026videochat3}.
Our token-level loss masking retains all \texttt{Response} targets while subsampling repetitive \texttt{Silence} and \texttt{Hold} continuations. Chunks containing only \texttt{Silent} responses also undergo masking.
The train-time sequence also appends the natural language response after each response token, where the language tokens undergo no loss masking.

Overall, our inference harness and SFT pipeline share a single streaming interface, and demonstrate how offline VLMs can be adapted to streaming inference eliciting their proactive responsiveness.

\section{Experiments}
\label{sec:experiments}

\paragraph{Experimental Setup.}
We train on the \dataset train split and select harnesses and checkpoints by validation PAUC-F; the four test benchmarks are held out.
Overlap filtering is described in \Cref{app:dataset_documentation}.
We report PAUC-F (GPT-4.1 judge) on ProactiveVideoQA and ESTP-Bench Explicit Proactive.
Configurations are in \Cref{app:harness,app:training}.
Both training and inference use $8\times$ H200 GPU nodes. 

\begin{table}[t]
\caption{
\textbf{Results on ProactiveVideoQA Benchmark.}
We report PAUC-F with varying $\omega$ using GPT-4.1 judge. 
Harness is ours for offline models and default from streaming native models.
Our harness is modified to integrate with Gemini-Live (ours-v) due to closed API nature.
\textbf{Best} and \underline{second-best} marked in each column.
}
\vspace{-0.5em}
\label{tab:proactive_videoqa}
\centering
\small
\def\arraystretch{1.3}  
\setlength\tabcolsep{0.7em}  
\scalebox{0.60}{%
\begin{tabular}{lc|ccc|ccc|ccc|ccc|ccc}
    \toprule
    Model & Harness & \multicolumn{3}{c|}{\texttt{[WEB]}} & \multicolumn{3}{c|}{\texttt{[EGO]}} & \multicolumn{3}{c|}{\texttt{[TV]}} & \multicolumn{3}{c|}{\texttt{[VAD]}} & \multicolumn{3}{c}{\texttt{[Mean]}} \\
    \hline
    $\omega=$ & & 0.0 & 0.5 & 1.0 & 0.0 & 0.5 & 1.0 & 0.0 & 0.5 & 1.0 & 0.0 & 0.5 & 1.0 & 0.0 & 0.5 & 1.0 \\
    \hline
    \hline
    \multicolumn{17}{l}{\textit{Proprietary Models}} \\
    \hline
    GPT-4.1 & ours & \textbf{57.92} & \underline{60.19} & \underline{62.31} & 33.15 & 34.33 & 35.36 & 31.30 & 32.56 & 33.58 & 28.00 & 28.80 & 29.49 & 37.59 & 38.97 & 40.19 \\
    GPT-4.1-mini & ours & 54.36 & 57.19 & 59.81 & 21.63 & 22.07 & 22.44 & 24.07 & 24.82 & 25.42 & 19.92 & 20.36 & 20.74 & 30.00 & 31.11 & 32.10 \\
    GPT-5.6-Sol & ours & \underline{57.62} & \textbf{60.35} & \textbf{62.87} & 31.01 & 32.03 & 32.92 & \underline{38.75} & \underline{40.94} & \underline{42.74} & 33.59 & 34.81 & 35.90 & 40.24 & 42.03 & 43.61 \\
    Gemini-2.5-Pro & ours & 55.13 & 57.67 & 60.01 & 32.70 & 33.97 & 35.06 & 36.14 & 37.98 & 39.48 & 23.84 & 24.53 & 25.11 & 36.95 & 38.54 & 39.92 \\
    Gemini-2.5-Flash & ours & 40.02 & 41.31 & 42.45 & 9.27 & 9.33 & 9.38 & 17.92 & 18.33 & 18.65 & 4.31 & 4.33 & 4.35 & 17.88 & 18.33 & 18.71 \\
    Gemini-3.1-Pro & ours & 50.21 & 52.76 & 55.00 & 19.83 & 20.13 & 20.38 & 37.62 & 39.49 & 41.02 & 0.66 & 0.66 & 0.66 & 27.08 & 28.26 & 29.27 \\
    Gemini-3.5-Flash-Lite & ours & 48.78 & 51.20 & 53.33 & 11.14 & 11.23 & 11.31 & 20.44 & 20.96 & 21.38 & 0.44 & 0.44 & 0.44 & 20.20 & 20.96 & 21.62 \\
    Gemini-3.8-Flash & ours & 51.22 & 53.84 & 56.15 & 19.12 & 19.44 & 19.71 & 30.56 & 31.78 & 32.76 & 0.33 & 0.33 & 0.33 & 25.31 & 26.35 & 27.24 \\
    Gemini-Live-2.5 & default & 27.56 & 27.56 & 27.56 & 0.00 & 0.00 & 0.00 & 15.48 & 15.48 & 15.48 & 0.00 & 0.00 & 0.00 & 10.76 & 10.76 & 10.76 \\
    Gemini-Live-2.5 & ours-v & 42.04 & 46.59 & 50.29 & 9.89 & 10.00 & 10.09 & 6.85 & 6.94 & 7.03 & 7.60 & 7.65 & 7.69 & 16.60 & 17.80 & 18.78 \\

    \hline
    \multicolumn{17}{l}{\textit{Open-Weight Models (No Streaming Video Training)}} \\
    \hline
    Molmo2 8B & ours & 36.17 & 38.22 & 40.07 & 10.70 & 10.81 & 10.90 & 9.79 & 9.88 & 9.95 & 6.61 & 6.64 & 6.66 & 15.82 & 16.39 & 16.90 \\
    MiniCPM-V 4.5 & ours & 37.16 & 39.43 & 41.40 & 11.88 & 12.04 & 12.18 & 4.38 & 4.40 & 4.41 & 4.38 & 4.39 & 4.40 & 14.45 & 15.07 & 15.60 \\
    Inkling-Small & ours & 48.32 & 50.42 & 52.39 & 24.62 & 25.39 & 26.06 & 15.11 & 15.31 & 15.49 & 9.62 & 9.66 & 9.69 & 24.42 & 25.20 & 25.91 \\
    Gemma4-A4B & ours & 41.92 & 43.39 & 44.71 & 34.63 & 35.94 & 37.15 & 26.17 & 26.60 & 27.01 & 27.82 & 28.50 & 29.13 & 32.64 & 33.61 & 34.50 \\
    Qwen3.5-A3B & ours & 51.85 & 54.24 & 56.54 & 37.62 & 39.29 & 40.82 & 33.42 & 34.47 & 35.46 & \underline{41.90} & \underline{43.75} & \underline{45.51} & \underline{41.20} & \underline{42.94} & \underline{44.58} \\
    \hline
    \multicolumn{17}{l}{\textit{Streaming Video Models (Trained for Streaming Responses)}} \\
    \hline
    VideoLLM-Online & default & 32.19 & 32.20 & 32.20 & 6.64 & 6.64 & 6.64 & 2.45 & 2.45 & 2.46 & 0.00 & 0.00 & 0.00 & 10.32 & 10.32 & 10.33 \\
    EyeWO & default & 38.97 & 40.42 & 41.81 & 4.76 & 4.77 & 4.78 & 3.24 & 3.23 & 3.23 & 0.00 & 0.00 & 0.00 & 11.74 & 12.11 & 12.46 \\
    MMDuet & default & 48.43 & 50.76 & 52.92 & 7.48 & 7.52 & 7.55 & 2.70 & 2.71 & 2.72 & 2.41 & 2.42 & 2.42 & 15.26 & 15.85 & 16.40 \\
    MMDuet2 & default & 50.69 & 56.51 & 61.46 & 31.64 & 32.28 & 32.89 & 30.34 & 32.29 & 33.91 & 14.70 & 14.97 & 15.22 & 31.84 & 34.01 & 35.87 \\
    VideoChat3 & default & 33.68 & 34.70 & 35.68 & 28.86 & 29.71 & 30.51 & 36.90 & 39.53 & 41.89 & 22.84 & 23.26 & 23.66 & 30.57 & 31.80 & 32.94 \\ \rowcolor{techgray}
    \gemmaModel (ours) & ours & 52.19 & 56.46 & 60.38 & \underline{41.81} & \underline{43.80} & \underline{45.61} & 31.65 & 33.27 & 34.70 & 30.85 & 31.98 & 33.01 & 39.13 & 41.38 & 43.43 \\ \rowcolor{techgray}
    \qwenModel (ours) & ours & 53.13 & 57.48 & 61.41 & \textbf{43.77} & \textbf{46.05} & \textbf{48.13} & \textbf{40.71} & \textbf{43.11} & \textbf{45.25} & \textbf{46.56} & \textbf{49.67} & \textbf{52.55} & \textbf{46.04} & \textbf{49.08} & \textbf{51.84} \\

    \bottomrule
\end{tabular}
}\end{table}

\begin{table}[t]
\caption{
\textbf{Results on ESTP-Bench.}
We report PAUC-F ($\omega$ = 0.5, GPT-4.1 judge) on Explicit Proactive subset of ESTP-Bench.
Our harness used for offline models and default harness for streaming models. 
The task categories are Object Recognition (OR), Attribute Perception (AP), Text-Rich Understanding (TRU), Object Localization (OL), Object State Change Recognition (OSC), Ego Object Localization (EOL), Ego Object State Change Recognition (EOSC), and Action Recognition (AR). 
\textbf{Best} and \underline{second-best} marked in each column. 
}
\label{tab:estp_paucf}
\vspace{-0.5em}
\centering
\small
\def\arraystretch{1.3}
\setlength\tabcolsep{1.2em}
\scalebox{0.7}{%
\begin{tabular}{lc|ccccccccc}
    \toprule
    Model & Harness & OR & AP & TRU & OL & OSC & EOL & EOSC & AR & All \\
    \hline
    \hline
    \multicolumn{11}{l}{\textit{Proprietary Models}} \\
    \hline
    GPT-4.1 & ours & 20.12 & 26.88 & 23.57 & 23.60 & 17.89 & 19.71 & 19.51 & \underline{24.01} & 21.91 \\
    GPT-4.1-mini & ours & 21.80 & \underline{27.41} & 27.04 & \textbf{31.39} & 19.48 & \underline{26.51} & 18.17 & 21.81 & 24.20 \\
    Gemini-2.5-Pro & ours & \textbf{25.36} & 27.14 & \textbf{34.53} & \underline{30.14} & 21.01 & 19.36 & 18.94 & 22.89 & 24.92 \\
    Gemini-2.5-Flash & ours & 9.44 & 15.67 & 19.02 & 9.40 & 9.22 & 8.04 & 7.62 & 12.98 & 11.42 \\
    \hline
    \multicolumn{11}{l}{\textit{Open-Weight Models (No Streaming Video Training)}} \\
    \hline
    Molmo2 8B & ours & 8.84 & 11.18 & 13.01 & 4.93 & 3.98 & 5.57 & 3.25 & 8.04 & 7.35 \\
    MiniCPM-V 4.5 & ours & 6.75 & 9.35 & 3.20 & 8.10 & 4.11 & 6.32 & 3.98 & 6.38 & 6.02 \\
    Inkling-Small & ours & 10.28 & 14.11 & 10.45 & 13.27 & 9.08 & 10.91 & 8.12 & 12.25 & 11.06 \\
    Gemma4-A4B & ours & 11.05 & 16.90 & 15.59 & 14.07 & 7.52 & 11.16 & 6.41 & 10.33 & 11.63 \\
    Qwen3.5-A3B & ours & 23.88 & 27.17 & 23.07 & 29.78 & \underline{29.24} & 23.10 & \underline{29.87} & 23.47 & \underline{26.20} \\
    \hline
    \multicolumn{11}{l}{\textit{Streaming Video Models (Trained for Streaming Responses)}} \\
    \hline
    VideoLLM-Online & default & 0.73 & 0.08 & 0.05 & 0.33 & 1.19 & 0.45 & 0.45 & 2.53 & 0.73 \\
    EyeWO & default & 10.86 & 7.08 & 3.43 & 10.72 & 7.16 & 10.20 & 4.56 & 11.23 & 8.16 \\
    MMDuet & default & 1.91 & 2.96 & 0.72 & 3.24 & 3.83 & 3.06 & 2.43 & 6.22 & 3.05 \\
    MMDuet2 & default & 15.47 & 20.89 & 14.15 & 14.74 & 11.32 & 11.50 & 10.43 & 10.17 & 13.58 \\
    VideoChat3 & default & 12.63 & 16.71 & 13.07 & 19.83 & 12.08 & 17.19 & 12.76 & 11.65 & 14.49 \\ \rowcolor{techgray}
    \gemmaModel (ours) & ours & 20.23 & 25.97 & 30.06 & 29.83 & 20.89 & 24.41 & 17.73 & 21.28 & 23.80 \\ \rowcolor{techgray}
    \qwenModel (ours) & ours & \underline{24.62} & \textbf{31.77} & \underline{30.47} & 29.08 & \textbf{30.82} & \textbf{27.84} & \textbf{31.64} & \textbf{24.24} & \textbf{28.81} \\
    \bottomrule
\end{tabular}%
}\end{table}

\subsection{Main Results}

\paragraph{ProactiveVideoQA.}
We present PAUC-F evaluations across the four splits from \citet{wang2025proactivevqa} in \Cref{tab:proactive_videoqa}. 
All offline proprietary and open-weight models are evaluated using our harness with per-model harness calibration. 
The Gemini-Live model is evaluated with its default harness as well as an integration of our harness (\texttt{ours-v}) without calibration. 
For streaming video model baselines, we evaluate the official model checkpoints with their default streaming inference settings. 
Our two trained models, \gemmaModel and \qwenModel, use our harness for inference. 
Our trained \qwenModel model achieves state-of-the-art performance overall, demonstrating the usefulness of our \framework framework. \qwenModel (49.08) also outperforms prior best MMDuet2 (34.01) by 44\%.
We also report the raw PAUC numbers (official metric of benchmark) in \Cref{app:additional_experiments} for reference.

\begin{table*}[t]
\centering
\footnotesize
\captionsetup{skip=4pt,font=scriptsize}
\begin{minipage}[t]{0.39\textwidth}
\centering
\caption{
ESTP-F1 on ESTP-Bench with baseline results identical to \citet{zhang2025eyeswideopenego}.
Our methods (both harness and trained) are highlighted in gray.
\textbf{Best} and \underline{second-best} marked in each column.
}
\label{tab:ESTP_result}
\def\arraystretch{1.2}
\setlength{\tabcolsep}{8pt}
\scalebox{0.70}{%
\begin{tabular}{lcccc}
\toprule
Model & Exp. & Imp. & Ctx. & All \\
\midrule
\multicolumn{5}{l}{\textit{Offline MLLMs}} \\
\midrule
LLaVA-OV & 13.6 & 31.8 & 8.2 & 18.0 \\
Qwen2-VL & 15.4 & \underline{39.3} & 10.4 & 21.3 \\
MiniCPM-V & 15.2 & 36.5 & 26.6 & 22.9 \\
LLaVA-NeXT & 16.5 & 34.6 & 13.8 & 21.3 \\
InternVL-V2 & 5.6 & 6.7 & 5.7 & 5.9 \\
\midrule
\multicolumn{5}{l}{\textit{Online MLLMs}} \\
\midrule
LIVE$_{0.8}$ & 7.0 & 9.3 & 11.8 & 9.1 \\
LIVE$_{0.9}$ & 9.5 & 25.6 & 18.9 & 15.5 \\
MMDuet & 9.1 & 34.2 & 20.3 & 17.8 \\
EyeWO & 23.6 & \textbf{52.5} & \textbf{43.6} & \underline{34.7} \\ \rowcolor{techgray}
Gemma4-A4B & 18.2 & 31.2 & 27.1 & 23.2 \\ \rowcolor{techgray}
Qwen3.5-A3B & 16.6 & 22.6 & 14.1 & 18.0 \\ \rowcolor{techgray}
\gemmaModel & \underline{31.6} & 30.4 & \underline{40.5} & 32.5 \\ \rowcolor{techgray}
\qwenModel & \textbf{35.2} & 32.1 & 40.4 & \textbf{35.1} \\
\bottomrule
\end{tabular}%
}
\end{minipage}
\hfill
\begin{minipage}[t]{0.32\textwidth}
\centering
\caption{
StreamingBench Proactive Split (PO) time accuracy (follow official leaderboard) \& our interactive protocol on same split (PO-I).
\textbf{Best} and \underline{second-best} marked in each column.
}
\label{tab:sb_po}
\def\arraystretch{1.25}
\setlength{\tabcolsep}{8pt}
\scalebox{0.70}{%
\begin{tabular}{lcc}
\toprule
Model & PO & PO-I \\
\midrule
Gemini 1.5 Pro & \underline{45.10} & -- \\
GPT-4o & 39.22 & -- \\
Claude 3.5 Sonnet & 35.29 & -- \\ \midrule
VideoLLM-Online & 1.96 & -- \\
VITA-1.5 & 25.49 & -- \\
Dispider & 25.34 & -- \\
InternVL2 & 20.45 & -- \\
MiniCPM-o 2.6 & 29.41 & 30.27 \\
MMDuet2 & 34.69 & 5.95 \\
VideoChat3 & 9.80 & 18.38 \\
\rowcolor{techgray}
Gemma4-A4B & 36.80 & 38.9 \\ \rowcolor{techgray}
Qwen3.5-A3B & 25.20 & 15.1 \\ \rowcolor{techgray}
\gemmaModel & 44.00 & \underline{48.6} \\ \rowcolor{techgray}
\qwenModel & \textbf{47.60} & \textbf{54.6} \\
\bottomrule
\end{tabular}%
}
\end{minipage}\hfill
\begin{minipage}[t]{0.24\textwidth}
\centering
\caption{
OVO-Bench Forward CRR Task. Baseline results follow \citet{li2025ovobench}.
Our methods highlighted gray.
\textbf{Best} and \underline{second-best} marked.
}
\label{tab:ovo_forward}
\def\arraystretch{1.4}
\setlength{\tabcolsep}{8pt}
\scalebox{0.70}{%
\begin{tabular}{lc}
\toprule
Model & CRR \\
\midrule
Gemini 1.5 Pro & \underline{61.67} \\
GPT-4o & 59.40 \\ \midrule
LLaVA-Video-7B & 60.42 \\
LLaVA-OV-7B & 58.75 \\
Qwen2-VL-72B & 45.00 \\
Qwen2-VL-7B & 48.75 \\ \rowcolor{techgray}
Gemma4-A4B & 59.58 \\ \rowcolor{techgray}
Qwen3.5-A3B & 56.67 \\ \rowcolor{techgray}
\gemmaModel & 60.83 \\ \rowcolor{techgray}
\qwenModel & \textbf{62.50} \\
\bottomrule
\end{tabular}%
}
\end{minipage}
\vspace{-0.5em}
\end{table*}

\paragraph{ESTP-Bench.}
We next report PAUC-F on the \textit{Explicit Proactive} subset of tasks from \citet{zhang2025eyeswideopenego}, which measures proactive responsiveness. 
Our results in \Cref{tab:estp_paucf} indicate state-of-the-art performance of proposed \gemmaModel and \qwenModel.
All baselines use the same official model checkpoints with our harness, with streaming native models using their default inference setting.

\paragraph{Official ESTP-Bench, StreamingBench (PO), and OVOBench (CRR).}
We next report results on existing benchmarks using their default metrics, allowing direct comparison to baseline results originally reported in \citet{zhang2025eyeswideopenego,lin2024streamingbench,li2025ovobench}. 
We report results for Gemma4 and Qwen3.5 offline models converted to online streaming setting with our harness, as well as \gemmaModel and \qwenModel which are our two trained models.

In \Cref{tab:ESTP_result}, we report ESTP-F1 metric on full ESTP-Bench benchmark. 
Baseline results are identical to \citet{zhang2025eyeswideopenego}. 
Our \qwenModel model achieves best overall performance and on Explicit (Exp.) split. 
Implicit (Imp.) and Contextual (Ctx.) splits contain longer gold intervals and because ESTP-F1 overlooks in-window repetition, models with repetitive behavior can score higher on these (more in \Cref{tab:estp_split_windows}).

In \Cref{tab:sb_po}, we report time accuracy for the Proactive (PO) split of StreamingBench with baseline results obtained from official leaderboard. 
Our \qwenModel model similarly achieves overall best performance. 
We additionally introduce an interactive split (PO-I) where new queries are introduced mid-video in a true streaming fashion (more details in \Cref{app:po_interactive}). Baselines are evaluated using official checkpoints under identical settings as our models. 

In \Cref{tab:ovo_forward}, we report accuracy for the CRR (Clues Reveal Responding) task of OVOBench. 
In this task, a query is provided before the video contains sufficient information to respond, and models are expected to wait until evidence appears and then proactively respond. 
This is the key task in OVOBench that measures proactive responsiveness; hence we use it in our evaluations. 
In this task too, our \qwenModel model achieves overall best performance.

\begin{table*}[t]
\centering
\footnotesize
\captionsetup{skip=4pt,font=scriptsize}
\begin{minipage}[t]{0.32\linewidth}
\centering
\caption{
\textbf{Annotation \& Data Scale Ablation.}
PAUC-F ($\omega=0.5$) on EGO \& WEB for \gemmaModel.
Each stage of our dataset construction (human review and filtering) improves performance. 
Results also scale when training with increasing data. 
}
\label{tab:annotation_ablation}
\def\arraystretch{1.3}
\setlength{\tabcolsep}{0.6em}
\scalebox{0.60}{%
\begin{tabular}{@{}lccc@{}}
\toprule
Version & Frac. & WEB $\uparrow$ & EGO $\uparrow$ \\
\midrule
Raw Gemini-generated       & 100\% & 54.8 & 42.4\\
Intermediate, pre-cleaning & 100\% & 55.4 & 42.9 \\ \rowcolor{techgray}
Final human-verified       & 100\% & 56.5 & 43.8 \\
Final human-verified       & 10\%  & 53.8 & 42.3 \\
\bottomrule
\end{tabular}%
}
\end{minipage}
\hfill
\begin{minipage}[t]{0.32\linewidth}
\centering
\caption{
\textbf{Inference Harness Ablation.}
PAUC-F ($\omega=0.5$) on WEB and EGO for Gemma4-A4B.
Baseline from VideoChat3 for comparison purposes. 
}
\label{tab:inference_harness_ablation}
\def\arraystretch{1.3}
\setlength\tabcolsep{0.8em}
\scalebox{0.68}{%
\begin{tabular}{@{}lcc@{}}
\toprule
Method & WEB & EGO \\ \midrule \rowcolor{techgray}
Ours (Default)              & 43.4 & 36.0 \\
Ours w/o sliding window     & 33.6 &  7.0 \\
Ours w/o online memory      & 36.8 & 14.1 \\
Ours w/o structured output  & 42.2 & 34.6 \\
VideoChat3 Harness          & 26.3 &  2.4 \\
\bottomrule
\end{tabular}%
}
\end{minipage}
\hfill
\begin{minipage}[t]{0.32\linewidth}
\centering
\caption{
\textbf{Ablation on Training Pipeline.}
PAUC-F ($\omega=0.5$) on WEB \& EGO for \gemmaModel.
\texttt{Memory} and \texttt{Hold-State} help avoid repetitive firing. \texttt{Loss-Masking} is crucial to avoid collapsing to silence. 
}
\label{tab:training_ablation}
\def\arraystretch{1.3}
\setlength\tabcolsep{0.6em}
\scalebox{0.65}{%
\begin{tabular}{@{}ccccc@{}}
\toprule
Memory & \shortstack{Hold\\State} & \shortstack{Loss\\Masking} & WEB $\uparrow$ & EGO $\uparrow$ \\
\midrule
\rowcolor{techgray}
\ymark & \ymark & \ymark & 56.5 & 43.8 \\
\xmark & \ymark & \ymark & 54.8 & 40.5 \\
\ymark & \xmark & \ymark & 32.9 & 24.4 \\
\ymark & \ymark & \xmark & 0.0 & 0.0 \\
\bottomrule
\end{tabular}%
}
\end{minipage}
\vspace{-1.0em}
\end{table*}

\begin{table}[t]
\centering
\caption{\textbf{Controlled training comparisons on ProactiveVideoQA.} PAUC-F at $\omega=0.5$ with a common Qwen3.5-A3B backbone. Panel (a) fixes the \framework training recipe and inference harness while varying training data; panel (b) compares each method's own dataset and training recipe. Mean averages the four splits.}
\label{tab:controlled_training}
\vspace{-0.5em}
\small
\renewcommand{\arraystretch}{1.2}
\setlength{\tabcolsep}{1.1em}
\scalebox{0.85}{
\begin{tabular}{lrrrrr}
\toprule
Dataset / Method & WEB & EGO & TV & VAD & Mean \\
\midrule
\multicolumn{6}{l}{\textbf{(a) Training-data ablation: fixed backbone, recipe, and harness}} \\
Base model (no finetuning) & 54.24 & 39.29 & 34.47 & 43.75 & 42.94 \\
MMDuet2 data & 27.34 & 24.39 & 16.49 & 7.46 & 18.92 \\
VideoChat3 data & 55.15 & 39.71 & 40.53 & 45.72 & 45.28 \\
\rowcolor{techgray}
\dataset data & \textbf{57.48} & \textbf{46.05} & \textbf{43.11} & \textbf{49.67} & \textbf{49.08} \\
\midrule
\multicolumn{6}{l}{\textbf{(b) Training-method comparison: fixed backbone, original data and recipes}} \\
MMDuet2 & 32.76 & 29.57 & 22.20 & 13.03 & 24.39 \\
VideoChat3 & 35.86 & 30.36 & 40.75 & 24.67 & 32.91 \\
\rowcolor{techgray}
\qwenModel & \textbf{57.48} & \textbf{46.05} & \textbf{43.11} & \textbf{49.67} & \textbf{49.08} \\
\bottomrule
\end{tabular}}
\end{table}

\subsection{Ablative Studies}

\paragraph{Dataset.}
We establish the contribution of our human review and filtering during dataset construction in \Cref{tab:annotation_ablation}. 
Results also highlight how performance scales going from 10\% to 100\% of data in \dataset for model training. 

\begin{table}[ht!]
\centering
\caption{Hold-state diagnostics for \gemmaModel on ProactiveVideoQA ($\omega=0.5$). Repeated fires are responses repeating previously reported content, expressed as a percentage of all fires. Removing Hold increases firing and repetition while PAUC remains approximately unchanged.}
\label{tab:hold_diagnostics}
\vspace{-0.5em}
\small
\renewcommand{\arraystretch}{1.2}
\setlength{\tabcolsep}{5pt}
\scalebox{0.85}{
\begin{tabular}{llrrrrr}
\toprule
Split & Hold & PAUC-F & Fires & Precision (\%) & PAUC (\%) & Repeated (\%) \\
\midrule
WEB & With & 56.46 & 978 & 75.1 & 45.26 & 15 \\
WEB & Without & 32.90 & 2,840 & 25.8 & 45.39 & 65 \\
EGO & With & 43.80 & 1,955 & 42.2 & 45.52 & 40 \\
EGO & Without & 24.40 & 4,950 & 16.7 & 45.28 & 75 \\
\bottomrule
\end{tabular}}
\end{table}

\paragraph{Training Data.}
\Cref{tab:controlled_training}(a) fixes the Qwen3.5-A3B backbone, our training recipe, and inference harness, varying only the training dataset. \dataset yields the best performance across all four splits. Finetuning on \dataset improves mean PAUC-F from 42.94 to 49.08, a 6.14-point gain over the same pretrained model with our harness.

\paragraph{Training Method.}
\Cref{tab:controlled_training}(b) holds the backbone fixed while MMDuet2 and VideoChat3 retain their original datasets and training recipes. \qwenModel outperforms both, separating the combined contribution of our data and training pipeline from backbone differences. Full results at all three $\omega$ values and target-conversion details are in \Cref{app:additional_experiments}.

\paragraph{Inference Harness.}
In \Cref{tab:inference_harness_ablation}, we first ablate the key components of our proposed inference harness to verify their usefulness. 
Next we compare against an existing generic harness design from \citet{li2026videochat3} to demonstrate our gains in proactive responsiveness. 

\paragraph{Training Pipeline.}
We ablate our training pipeline in \Cref{tab:training_ablation}. 
Memory and Hold discourage repetitive firing. On WEB, removing Hold increases fires from 978 to 2,840 and lowers firing precision from 75.1\% to 25.8\%, while PAUC remains approximately unchanged (45.26\% vs.\ 45.39\%). This supports improved firing selectivity as the mechanism; full WEB/EGO diagnostics are in \Cref{tab:hold_diagnostics}. 
Loss masking is crucial to prevent models from collapsing to always remaining silent, as query-relevant events are sparse and the decision-token distribution is heavily skewed toward silence.

\section{Conclusion}

Proactive streaming video agents must decide between responding and remaining silent repeatedly for each timestep.
We formulate this task as a series of causal response decisions at each timestep. 
Our \framework framework unifies data, metric, harness, and training around this formulation and demonstrates state-of-the-art performance across four proactive streaming benchmarks.


\subsection*{AI use statement}

In this work, we used generative AI tools for following tasks: generate synthetic data sets, implement methods, clean and reformat dataset, propose or refine hypotheses, design or provide feedback on research methodology or experiments.
To reiterate, as explained in the paper, generative AI tools were used for automated preliminary dataset creation (part of our contribution) and automated inference harness tuning (part of our contribution).
We have not used generative AI tools for the following tasks:
help develop theoretical models or conceptual frameworks, formulate mathematical claims, provide critical ingredients for proving mathematical claims, assist in the writing of proofs, assist with translation, support qualitative and thematic data analysis, interpret results. 
The rest of the required disclosure tasks are not applicable to this work.
Additionally, we used generative AI tools for the following:
formulate questions for surveys or interviews, 
create or modify scientific figures or images, 
create or edit software code, 
summarize or analyse existing literature, 
sourcing/searching for information, 
identify relevant literature, 
format references, 
propose a title.
We have reviewed all AI-assisted work. 
Particularly, all AI generated code and AI assisted literature survey have been verified by authors.
We take responsibility for the final content of this work, including text, claims or artifacts produced with the aid of generative AI.


\subsection*{Reproducibility statement}
Our data, code, and trained model checkpoints will be released publicly. 
For closed source models, all generation outputs will be released. 
For open models, we use publicly available checkpoints in all our evaluations.

\subsection*{Author Contributions}
Kanchana Ranasinghe led the project, developing the inference harness, PAUC-F metric, and training pipelines, and running most of the training, inference, and evaluation experiments.

Honglu Zhou set the research and experimental directions, led dataset annotation, conducted initial training and metric-comparison experiments, and contributed substantially to the project.

Dominick Reilly helped run baselines, surveyed the literature, and gave feedback on the method and the manuscript.

Le Xue and Shusheng Yang built the Gemini question-generation pipeline used in dataset creation; Le Xue also contributed to discussions of the dataset design.

Juan Carlos Niebles, Silvio Savarese, and Ran Xu provided high-level direction and discussed many aspects of the project.

\subsection*{Acknowledgments}
We thank Artemis Panagopoulou for reviewing this paper and providing extensive feedback.
We also thank the broader team at Salesforce AI Research for their support.

\bibliographystyle{iclr2027_conference}
\bibliography{main}

@misc{wang2025proactivevqa,
  title={ProactiveVideoQA: A Comprehensive Benchmark Evaluating Proactive Interactions in Video Large Language Models},
  author={Yueqian Wang and Xiaojun Meng and Yifan Wang and Huishuai Zhang and Dongyan Zhao},
  year={2025},
  eprint={2507.09313},
  archivePrefix={arXiv},
  primaryClass={cs.CV},
}

@misc{yao2026harnessing,
  title={Harnessing Streaming Video in the Wild},
  author={Dingyu Yao and Shuhuan Gu and Qingyi Si and Junhao Zhou and Chenxu Yang and Chuanyu Qin and Naibin Gu and Zheng Lin and Weiping Wang and Nan Duan and Jiaqi Wang},
  year={2026},
  eprint={2606.08615},
  archivePrefix={arXiv},
  primaryClass={cs.CV},
}

@inproceedings{zhang2025eyeswideopenego,
  title={Eyes Wide Open: Ego Proactive {Video-LLM} for Streaming Video},
  author={Xueyang Yu and Cheng Shi and Yang Wang and Sibei Yang},
  booktitle={Advances in Neural Information Processing Systems},
  year={2025},
  doi={10.52202/085713-0453},
}

@inproceedings{wang2024mmduet,
  title={VideoLLM Knows When to Speak: Enhancing Time-Sensitive Video Comprehension with Video-Text Duet Interaction Format},
  author={Yueqian Wang and Xiaojun Meng and Yuxuan Wang and Jianxin Liang and Jiansheng Wei and Huishuai Zhang and Dongyan Zhao},
  booktitle={Findings of the Association for Computational Linguistics: EMNLP 2025},
  pages={6338--6359},
  year={2025},
  publisher={Association for Computational Linguistics},
  doi={10.18653/v1/2025.findings-emnlp.336},
}

@inproceedings{wang2026mmduet2,
  title={MMDuet2: Enhancing Proactive Interaction of Video MLLMs with Multi-Turn Reinforcement Learning},
  author={Yueqian Wang and Songxiang Liu and Disong Wang and Nuo Xu and Guanglu Wan and Huishuai Zhang and Dongyan Zhao},
  booktitle={International Conference on Learning Representations},
  year={2026},
  eprint={2512.06810},
  archivePrefix={arXiv},
  primaryClass={cs.CV},
}

@inproceedings{chen2024videollmonline,
  title={VideoLLM-online: Online Video Large Language Model for Streaming Video},
  author={Joya Chen and Zhaoyang Lv and Shiwei Wu and Kevin Qinghong Lin and Chenan Song and Difei Gao and Jia-Wei Liu and Ziteng Gao and Dongxing Mao and Mike Zheng Shou},
  booktitle={Proceedings of the IEEE/CVF Conference on Computer Vision and Pattern Recognition},
  pages={18407--18418},
  year={2024},
}

@misc{li2026videochat3,
  title={VideoChat3: Fully Open Video MLLM for Efficient and Generalist Video Understanding},
  author={Xinhao Li and Yuhan Zhu and Xiangyu Zeng and Yuhao Dong and Haoning Wu and Zhiqiu Zhang and Yuandong Yang and Changlian Ma and Qingyu Zhang and Yansong Shi and Xinyu Chen and Haoran Chen and Zizheng Huang and Jun Zhang and Kun Ouyang and Lin Sui and Ziang Yan and Yicheng Xu and Chenting Wang and Yinan He and Hongjie Zhang and Yi Wang and Yu Qiao and Yali Wang and Ziwei Liu and Kai Chen and Limin Wang},
  year={2026},
  eprint={2607.14935},
  archivePrefix={arXiv},
  primaryClass={cs.CV},
}

@inproceedings{wang2025streambridge,
  title={{StreamBridge}: Turning Your Offline Video Large Language Model into a Proactive Streaming Assistant},
  author={Haibo Wang and Bo Feng and Zhengfeng Lai and Mingze Xu and Shiyu Li and Weifeng Ge and Afshin Dehghan and Meng Cao and Ping Huang},
  booktitle={Advances in Neural Information Processing Systems},
  volume={38},
  year={2025},
  doi={10.52202/085713-4406},
}

@inproceedings{qian2025dispider,
  title={{Dispider}: Enabling Video {LLMs} with Active Real-Time Interaction via Disentangled Perception, Decision, and Reaction},
  author={Rui Qian and Shuangrui Ding and Xiaoyi Dong and Pan Zhang and Yuhang Zang and Yuhang Cao and Dahua Lin and Jiaqi Wang},
  booktitle={Proceedings of the IEEE/CVF Conference on Computer Vision and Pattern Recognition},
  pages={24045--24055},
  year={2025},
}

@inproceedings{ding2025streammind,
  title={{StreamMind}: Unlocking Full Frame Rate Streaming Video Dialogue through Event-Gated Cognition},
  author={Xin Ding and Hao Wu and Yifan Yang and Shiqi Jiang and Qianxi Zhang and Donglin Bai and Zhibo Chen and Ting Cao},
  booktitle={Proceedings of the IEEE/CVF International Conference on Computer Vision},
  pages={13448--13459},
  year={2025},
}

@inproceedings{azad2026streamready,
  title={{StreamReady}: Learning What to Answer and When in Long Streaming Videos},
  author={Shehreen Azad and Vibhav Vineet and Yogesh S. Rawat},
  booktitle={Proceedings of the IEEE/CVF Conference on Computer Vision and Pattern Recognition},
  pages={40494--40504},
  year={2026},
}

@inproceedings{zhang2026progressive,
  title={Progressive Online Video Understanding with Evidence-Aligned Timing and Transparent Decisions},
  author={Kecheng Zhang and Zongxin Yang and Mingfei Han and Haihong Hao and Yunzhi Zhuge and Changlin Li and Junhan Zhao and Zhihui Li and Xiaojun Chang},
  booktitle={International Conference on Learning Representations},
  year={2026},
}

@misc{tian2026roma,
  title={{ROMA}: Real-Time Omni-Multimodal Assistant with Interactive Streaming Understanding},
  author={Xueyun Tian and Wei Li and Bingbing Xu and Heng Dong and Yuanzhuo Wang and Huawei Shen},
  year={2026},
  eprint={2601.10323},
  archivePrefix={arXiv},
  primaryClass={cs.CV},
}

@inproceedings{yao2025timechatonline,
  title={{TimeChat-Online}: 80\% Visual Tokens are Naturally Redundant in Streaming Videos},
  author={Linli Yao and Yicheng Li and Yuancheng Wei and Lei Li and Shuhuai Ren and Yuanxin Liu and Kun Ouyang and Lean Wang and Shicheng Li and Sida Li and Lingpeng Kong and Qi Liu and Yuanxing Zhang and Xu Sun},
  booktitle={Proceedings of the 33rd ACM International Conference on Multimedia},
  pages={10807--10816},
  year={2025},
}

@inproceedings{yang2025livestar,
  title={{LiveStar}: Live Streaming Assistant for Real-World Online Video Understanding},
  author={Zhenyu Yang and Kairui Zhang and Yuhang Hu and Bing Wang and Shengsheng Qian and Bin Wen and Fan Yang and Tingting Gao and Weiming Dong and Changsheng Xu},
  booktitle={Advances in Neural Information Processing Systems},
  volume={38},
  year={2025},
  doi={10.52202/085713-1050},
}

@article{zhang2024llava,
  title={Llava-video: Video instruction tuning with synthetic data},
  author={Zhang, Yuanhan and Wu, Jinming and Li, Wei and Li, Bo and Ma, Zejun and Liu, Ziwei and Li, Chunyuan},
  journal={arXiv preprint arXiv:2410.02713},
  year={2024}
}

@inproceedings{abreu2025parseego4d,
  title={{PARSE-Ego4D}: Toward Bidirectionally Aligned Action Recommendations for Egocentric Videos},
  author={Steven Abreu and Tiffany D. Do and Karan Ahuja and Eric J. Gonzalez and Lee Payne and Daniel McDuff and Mar Gonzalez-Franco},
  booktitle={ICLR Workshop on Bidirectional Human-AI Alignment},
  year={2025},
}

@inproceedings{xia2026streamingvideo,
  title={Streaming Video Instruction Tuning},
  author={Jiaer Xia and Peixian Chen and Mengdan Zhang and Xing Sun and Kaiyang Zhou},
  booktitle={Proceedings of the IEEE/CVF Conference on Computer Vision and Pattern Recognition},
  pages={31219--31229},
  year={2026},
}

@article{lin2024streamingbench,
  title={StreamingBench: Assessing the Gap for MLLMs to Achieve Streaming Video Understanding},
  author={Junming Lin and Zheng Fang and Chi Chen and Zihao Wan and Fuwen Luo and Peng Li and Yang Liu and Maosong Sun},
  journal={arXiv preprint arXiv:2411.03628},
  year={2024},
}

@inproceedings{li2025ovobench,
  title={OVO-Bench: How Far is Your Video-LLMs from Real-World Online Video Understanding?},
  author={Yifei Li and Junbo Niu and Ziyang Miao and Chunjiang Ge and Yuanhang Zhou and Qihao He and Xiaoyi Dong and Haodong Duan and Shuangrui Ding and Rui Qian and Pan Zhang and Yuhang Zang and Yuhang Cao and Conghui He and Jiaqi Wang},
  booktitle={Proceedings of the IEEE/CVF Conference on Computer Vision and Pattern Recognition},
  year={2025},
}

@inproceedings{zheng2026emgarde,
  title={Em-Garde: A Propose-Match Framework for Proactive Streaming Video Understanding},
  author={Zheng, Yikai and Ding, Xin and Yang, Yifan and Jiang, Shiqi and Wu, Hao and Zhang, Qianxi and Wang, Weijun and Cao, Ting and Liu, Yunxin},
  booktitle={Proceedings of the European Conference on Computer Vision (ECCV)},
  year={2026}
}

@misc{zhang2026darwinx,
  title={{DarwinX}: Evolving Agent Harnesses Through Natural Selection},
  author={Yifan Zhang and Yutong Dai and Juntao Tan and Luyu Yang and Rishi Mullur and Thai Hoang and Zhiyuan Hu and James Zhu and Phil Mui and Silvio Savarese and Ran Xu and Zeyuan Chen},
  year={2026},
  eprint={2608.07545},
  archivePrefix={arXiv},
  primaryClass={cs.NE},
}

@misc{xu2026videoharnessrsi,
  title={{VideoHarness-RSI}: Recursive Harness Self-Improvement for Long-Video Understanding with Frozen Vision-Language Models},
  author={Guoyang Xu and Hao Chen},
  year={2026},
  eprint={2608.24302},
  archivePrefix={arXiv},
  primaryClass={cs.AI},
}

@misc{zhang2026harnesscompass,
  title={{HarnessCompass}: Guiding Automatic Harness Evolution toward Generalizable and Effective Agent Harnesses},
  author={Luan Zhang and Ruochen Zhou and Dandan Song and Zhengyu Chen and Yuhang Tian and Jun Yang and Huipeng Ma and Chenhao Li and Guangyuan Feng and Xudong Li and Yizhou Jin and Yan Xu},
  year={2026},
  eprint={2608.01918},
  archivePrefix={arXiv},
  primaryClass={cs.LG},
}

@misc{gemmateam2026gemma4,
  title={{Gemma 4 Technical Report}},
  author={{Gemma Team}},
  year={2026},
  eprint={2607.02770},
  archivePrefix={arXiv},
  primaryClass={cs.CL},
}

@misc{qwenteam2026qwen35,
  title={{Qwen3.5-35B-A3B}},
  author={{Qwen Team}},
  year={2026},
  howpublished={Hugging Face model card},
}

@misc{clark2026molmo2,
  title={{Molmo2}: Open Weights and Data for Vision-Language Models with Video Understanding and Grounding},
  author={Christopher Clark and Jieyu Zhang and Zixian Ma and Jae Sung Park and Mohammadreza Salehi and Rohun Tripathi and Sangho Lee and Zhongzheng Ren and Chris Dongjoo Kim and Yinuo Yang and Vincent Shao and Yue Yang and Weikai Huang and Ziqi Gao and Taira Anderson and Jianrui Zhang and Jitesh Jain and George Stoica and Winson Han and Ali Farhadi and Ranjay Krishna},
  year={2026},
  eprint={2601.10611},
  archivePrefix={arXiv},
  primaryClass={cs.CV},
}

@inproceedings{yu2026minicpmv,
  title={{MiniCPM-V} 4.5: Cooking Efficient {MLLMs} via Architecture, Data, and Training Recipe},
  author={Tianyu Yu and Zefan Wang and Chongyi Wang and Fuwei Huang and Wenshuo Ma and Zhihui He and Tianchi Cai and Weize Chen and Yuxiang Huang and Yuanqian Zhao and Bokai Xu and Junbo Cui and Yingjing Xu and Liqing Ruan and Luoyuan Zhang and Hanyu Liu and Jingkun Tang and Hongyuan Liu and Qining Guo and Wenhao Hu and Bingxiang He and Jie Zhou and Jie Cai and Ji Qi and Zonghao Guo and Chi Chen and Guoyang Zeng and Yuxuan Li and Ganqu Cui and Ning Ding and Xu Han and Yuan Yao and Zhiyuan Liu and Maosong Sun},
  booktitle={Proceedings of the IEEE/CVF Conference on Computer Vision and Pattern Recognition},
  pages={11704--11715},
  year={2026},
}

@misc{thinkingmachines2026inkling,
  title={Inkling-Small},
  author={{Thinking Machines Lab}},
  year={2026},
  howpublished={\url{https://thinkingmachines.ai/news/inkling-small/}},
}

@misc{openbmb2025minicpmo,
  title={{MiniCPM-o} 2.6: A {GPT-4o} Level {MLLM} for Vision, Speech, and Multimodal Live Streaming on Your Phone},
  author={{OpenBMB}},
  year={2025},
  howpublished={Technical report},
  url={https://openbmb.notion.site/185ede1b7a558042b5d5e45e6b237da9},
}

@article{li2024llavaonevision,
  title={{LLaVA-OneVision}: Easy Visual Task Transfer},
  author={Bo Li and Yuanhan Zhang and Dong Guo and Renrui Zhang and Feng Li and Hao Zhang and Kaichen Zhang and Peiyuan Zhang and Yanwei Li and Ziwei Liu and Chunyuan Li},
  journal={arXiv preprint arXiv:2408.03326},
  year={2024},
}

@article{li2024llavanextinterleave,
  title={{LLaVA-NeXT-Interleave}: Tackling Multi-Image, Video, and {3D} in Large Multimodal Models},
  author={Feng Li and Renrui Zhang and Hao Zhang and Yuanhan Zhang and Bo Li and Wei Li and Zejun Ma and Chunyuan Li},
  journal={arXiv preprint arXiv:2407.07895},
  year={2024},
}

@article{wang2024qwen2vl,
  title={{Qwen2-VL}: Enhancing Vision-Language Model's Perception of the World at Any Resolution},
  author={Peng Wang and Shuai Bai and Sinan Tan and Shijie Wang and Zhihao Fan and Jinze Bai and Keqin Chen and Xuejing Liu and Jialin Wang and Wenbin Ge and Yang Fan and Kai Dang and Mengfei Du and Xuancheng Ren and Rui Men and Dayiheng Liu and Chang Zhou and Jingren Zhou and Junyang Lin},
  journal={arXiv preprint arXiv:2409.12191},
  year={2024},
}

@article{yao2024minicpm,
  title={{MiniCPM-V}: A {GPT-4V} Level {MLLM} on Your Phone},
  author={Yuan Yao and Tianyu Yu and Ao Zhang and Chongyi Wang and Junbo Cui and Hongji Zhu and Tianchi Cai and Haoyu Li and Weilin Zhao and Zhihui He and Qianyu Chen and Huarong Zhou and Zhensheng Zou and Haoye Zhang and Shengding Hu and Zhi Zheng and Jie Zhou and Jie Cai and Xu Han and Guoyang Zeng and Dahai Li and Zhiyuan Liu and Maosong Sun},
  journal={arXiv preprint arXiv:2408.01800},
  year={2024},
}

@article{chen2024internvl2,
  title={How Far Are We to {GPT-4V}? Closing the Gap to Commercial Multimodal Models with Open-Source Suites},
  author={Zhe Chen and Weiyun Wang and Hao Tian and Shenglong Ye and Zhangwei Gao and Erfei Cui and Wenwen Tong and Kongzhi Hu and Jiapeng Luo and Zheng Ma and Ji Ma and Jiaqi Wang and Xiaoyi Dong and Hang Yan and Hewei Guo and Conghui He and Botian Shi and Zhenjiang Jin and Chao Xu and Bin Wang and Xingjian Wei and Wei Li and Wenjian Zhang and Bo Zhang and Pinlong Cai and Licheng Wen and Xiangchao Yan and Min Dou and Lewei Lu and Xizhou Zhu and Tong Lu and Dahua Lin and Yu Qiao and Jifeng Dai and Wenhai Wang},
  journal={arXiv preprint arXiv:2404.16821},
  year={2024},
}

@inproceedings{radford2021clip,
  title={Learning Transferable Visual Models from Natural Language Supervision},
  author={Alec Radford and Jong Wook Kim and Chris Hallacy and Aditya Ramesh and Gabriel Goh and Sandhini Agarwal and Girish Sastry and Amanda Askell and Pamela Mishkin and Jack Clark and Gretchen Krueger and Ilya Sutskever},
  booktitle={International Conference on Machine Learning},
  pages={8748--8763},
  year={2021},
}

@inproceedings{zhao2023lavila,
  title={Learning Video Representations from Large Language Models},
  author={Yue Zhao and Ishan Misra and Philipp Kr{\"a}henb{\"u}hl and Rohit Girdhar},
  booktitle={Proceedings of the IEEE/CVF Conference on Computer Vision and Pattern Recognition},
  pages={6586--6597},
  year={2023},
}

@inproceedings{lin2022egovlp,
  title={Egocentric Video-Language Pretraining},
  author={Kevin Qinghong Lin and Jinpeng Wang and Mattia Soldan and Michael Wray and Rui Yan and Eric Zhongcong Xu and Difei Gao and Rongcheng Tu and Wenzhe Zhao and Weijie Kong and Chengfei Cai and Hongfa Wang and Dima Damen and Bernard Ghanem and Wei Liu and Mike Zheng Shou},
  booktitle={Advances in Neural Information Processing Systems},
  year={2022},
}

@article{fu2025vita,
  title={{VITA-1.5}: Towards {GPT-4o} Level Real-Time Vision and Speech Interaction},
  author={Chaoyou Fu and Haojia Lin and Xiong Wang and Yi-Fan Zhang and Yunhang Shen and Xiaoyu Liu and Yangze Li and Zuwei Long and Heting Gao and Ke Li and others},
  journal={arXiv preprint arXiv:2501.01957},
  year={2025},
}

@book{van1979information,
  title={Information Retrieval},
  author={Van Rijsbergen, C.J.},
  year={1979},
  publisher={Butterworths}
}

@book{yule1947introduction,
  title={An Introduction to the Theory of Statistics},
  author={Yule, G.U. and Kendall, M.G.},
  year={1947},
  publisher={C. Griffin, limited}
}

@article{Ferger1931TheNA,
  title={The Nature and Use of the Harmonic Mean},
  author={Wirth F. Ferger},
  journal={Journal of the American Statistical Association},
  year={1931}
}

\newpage
\appendix


\etocdepthtag.toc{appendix}
\addtocontents{toc}{\protect\setcounter{tocdepth}{1}}
\begingroup
\etocsettagdepth{main}{none}
\etocsettagdepth{appendix}{section}
\etocsettocstyle{\section*{Appendix Contents}}{}
\renewcommand{\cftsecaftersnum}{} 
\setlength{\cftsecnumwidth}{1.7em} 
\setlength{\cftbeforesecskip}{-0.1em} 
\hypersetup{linkcolor=black}        
\tableofcontents
\endgroup
\hypersetup{linkcolor=techblue}    

\renewcommand{\thetable}{A.\arabic{table}}
\renewcommand{\thefigure}{A.\arabic{figure}}
\setcounter{table}{0}
\setcounter{figure}{0}

\section{Additional Experiments}
\label{app:additional_experiments}

We report additional results proactive streaming video benchmarks, ProactiveVideoQA and ESTP-Bench as well as more splits of OVO-Bench. 
We also provide additional ablations comparing our method to existing harness designs and experiments involving our harness tuning. 

\paragraph{Baselines.}
We use official checkpoints of Molmo2 8B~\citep{clark2026molmo2}, MiniCPM-V 4.5~\citep{yu2026minicpmv}, Inkling-Small~\citep{thinkingmachines2026inkling}, MiniCPM-o 2.6~\citep{openbmb2025minicpmo}, VideoLLM-Online~\citep{chen2024videollmonline}, MMDuet~\citep{wang2024mmduet}, MMDuet2~\citep{wang2026mmduet2}, EyeWO~\citep{zhang2025eyeswideopenego}, and VideoChat3~\citep{li2026videochat3} for the baselines we report in our evaluations.
We also borrow numbers directly from prior evaluations in published papers for LLaVA-OneVision~\citep{li2024llavaonevision}, LLaVA-NeXT-Video~\citep{li2024llavanextinterleave}, LLaVA-Video~\citep{zhang2024llava}, Qwen2-VL~\citep{wang2024qwen2vl}, MiniCPM-V~\citep{yao2024minicpm}, InternVL2~\citep{chen2024internvl2}, CLIP~\citep{radford2021clip}, LaViLa~\citep{zhao2023lavila}, EgoVLP~\citep{lin2022egovlp}, VITA-1.5~\citep{fu2025vita}, and Dispider~\citep{qian2025dispider}.

\paragraph{ProactiveVideoQA PAUC.}
In \Cref{tab:proactive_videoqa_pauc}, we report the original PAUC metric \citep{wang2025proactivevqa} based evaluations for the same methods in \Cref{tab:proactive_videoqa}.
We highlight that the zero-point of PAUC is 25.0 (i.e. a model with no responses) and models scoring below 25 are worse than fully silent models (according to PAUC). 
Also note how models like Gemini-2.5-Flash (64.51) outperform the Gemini-2.5-Pro model (55.07) at $\omega=0.5$ in the Mean PAUC column, misaligning with our priors about these models' strengths. 
We reiterate our observations (see \Cref{fig:pauc_vs_paucf}) how models can game this metric with excessive firing, sacrificing precision for increased recall.  

Nevertheless, it is interesting to note that our models (\gemmaModel and \qwenModel) optimized for PAUC-F also achieve strong PAUC numbers while reducing excessive firing (see \Cref{tab:zss5_zss7_selection}), with \qwenModel achieving state-of-the-art performance among non-proprietary models.

\paragraph{ESTP-Bench.}
We also report the full per-task breakdown of ESTP-Bench ESTP-F1 (official metric) scores in \Cref{tab:ESTP_result_full}. 
These numbers correspond to the results summary in \Cref{tab:ESTP_result} of main paper.
Our \qwenModel showcases state-of-the-art overall performance on this benchmark as well. 
  
We note how on \textit{Implicit} and \textit{Contextual} splits several models outperform both our models by large margins. 
We argue this is due to weaknesses of the ESTP-F1 evaluation metric. 
Across the three splits, source videos are of similar length, but \textit{Implicit} and \textit{Contextual} gold windows are much longer (\Cref{tab:estp_split_windows}).
Models that repeat fires get penalized much less with the longer golden windows under ESTP-F1 metric. 
We highlight how our models outperform prior best EyeWO model across all 3 splits when evaluating with the more balanced PAUC-F metric in \Cref{tab:estp_paucf_full}.
\Cref{tab:estp_fires} shows why: EyeWO baseline emits an order of magnitude more fires per item than the ground-truth, especially on \textit{Implicit} and \textit{Contextual} splits, while \gemmaModel and \qwenModel stay near the gold rate.
Measuring repeats in this way to affirm prediction quality has been used in the prior work MMDuet2 \citep{wang2026mmduet2} as well. 

\begin{table}[t]
\caption{
\textbf{PAUC on ProactiveVideoQA Benchmark.}
We report PAUC with varying $\omega$ using GPT-4.1 judge, for the same runs, predictions, and judge configuration as \Cref{tab:proactive_videoqa}.
PAUC is the recall component of PAUC-F and does not penalize the number of responses emitted, so these values should be read together with \Cref{tab:proactive_videoqa}.
Models with excessive responsiveness (over-firing) can score higher on this metric than conservative models with balance precision and recall. 
Harness is ours for offline models and default for streaming native models.
Our harness is modified to integrate with Gemini-Live (ours-v) due to closed API nature.
\textbf{Best} and \underline{second-best} are marked separately within proprietary models and within the remaining models.
}
\vspace{-0.5em}
\label{tab:proactive_videoqa_pauc}
\centering
\small
\def\arraystretch{1.3}  
\setlength\tabcolsep{0.7em}  
\scalebox{0.60}{%
\begin{tabular}{lc|ccc|ccc|ccc|ccc|ccc}
    \toprule
    Model & Harness & \multicolumn{3}{c|}{\texttt{[WEB]}} & \multicolumn{3}{c|}{\texttt{[EGO]}} & \multicolumn{3}{c|}{\texttt{[TV]}} & \multicolumn{3}{c|}{\texttt{[VAD]}} & \multicolumn{3}{c}{\texttt{[Mean]}} \\
    \hline
    $\omega=$ & & 0.0 & 0.5 & 1.0 & 0.0 & 0.5 & 1.0 & 0.0 & 0.5 & 1.0 & 0.0 & 0.5 & 1.0 & 0.0 & 0.5 & 1.0 \\
    \hline
    \hline
    \multicolumn{17}{l}{\textit{Proprietary Models}} \\
    \hline
    GPT-4.1 & ours & 56.01 & 60.42 & 64.83 & 48.62 & 54.06 & 59.51 & 56.44 & 65.57 & 74.70 & \textbf{50.00} & \textbf{55.49} & \underline{60.98} & \underline{52.77} & \underline{58.89} & \underline{65.01} \\
    GPT-4.1-mini & ours & 47.22 & 51.67 & 56.12 & 54.44 & 60.49 & 66.54 & 55.31 & 64.25 & 73.20 & 45.27 & 50.20 & 55.14 & 50.56 & 56.65 & 62.75 \\
    GPT-5.6-Sol & ours & 52.54 & 57.26 & 61.97 & 49.44 & 55.04 & 60.63 & 52.15 & 60.90 & 69.65 & 43.28 & 47.58 & 51.87 & 49.35 & 55.19 & 61.03 \\
    Gemini-2.5-Pro & ours & 51.93 & 56.63 & 61.33 & 46.32 & 51.80 & 57.29 & 54.05 & 63.20 & 72.35 & 43.79 & 48.65 & 53.50 & 49.02 & 55.07 & 61.12 \\
    Gemini-2.5-Flash & ours & \textbf{60.62} & \textbf{67.00} & \textbf{73.38} & \underline{62.40} & \underline{68.78} & \underline{75.16} & \underline{57.61} & \underline{67.25} & \underline{76.90} & \underline{48.31} & \underline{55.00} & \textbf{61.68} & \textbf{57.23} & \textbf{64.51} & \textbf{71.78} \\
    Gemini-3.1-Pro & ours & 56.25 & 63.07 & 69.90 & 61.44 & 67.72 & 74.00 & 56.82 & 66.31 & 75.80 & 25.52 & 26.11 & 26.70 & 50.00 & 55.80 & 61.60 \\
    Gemini-3.5-Flash-Lite & ours & 55.83 & 62.59 & 69.35 & \textbf{64.67} & \textbf{71.57} & \textbf{78.48} & 53.29 & 61.15 & 69.00 & 23.42 & 23.28 & 23.13 & 49.30 & 54.65 & 59.99 \\
    Gemini-3.8-Flash & ours & \underline{56.80} & \underline{63.66} & \underline{70.52} & 60.37 & 67.32 & 74.27 & \textbf{58.29} & \textbf{68.32} & \textbf{78.35} & 25.13 & 25.35 & 25.57 & 50.15 & 56.16 & 62.18 \\
    Gemini-Live-2.5 & default & 24.10 & 24.10 & 24.10 & 25.00 & 25.00 & 25.00 & 20.31 & 20.31 & 20.31 & 25.00 & 25.00 & 25.00 & 23.60 & 23.60 & 23.60 \\
    Gemini-Live-2.5 & ours-v & 41.06 & 50.74 & 60.42 & 46.66 & 52.15 & 57.64 & 16.42 & 17.58 & 18.75 & 47.91 & 52.08 & 56.25 & 38.01 & 43.14 & 48.27 \\

    \hline
    \multicolumn{17}{l}{\textit{Open-Weight Models (No Streaming Video Training)}} \\
    \hline
    Molmo2 8B & ours & 34.80 & 38.80 & 42.81 & 35.39 & 37.81 & 40.24 & 28.70 & 30.23 & 31.75 & 26.99 & 27.86 & 28.74 & 31.47 & 33.68 & 35.88 \\
    MiniCPM-V 4.5 & ours & 40.43 & 46.22 & 52.01 & 39.07 & 42.87 & 46.67 & 23.11 & 23.83 & 24.55 & 19.40 & 19.75 & 20.09 & 30.50 & 33.17 & 35.83 \\
    Inkling-Small & ours & 40.87 & 43.96 & 47.04 & \textbf{43.31} & \textbf{48.54} & \textbf{53.76} & 32.84 & 34.77 & 36.70 & 30.59 & 31.30 & 32.01 & 36.90 & 39.64 & 42.38 \\
    Gemma4-A4B & ours & \textbf{52.45} & \textbf{57.30} & \textbf{62.14} & 34.81 & 37.57 & 40.32 & 28.70 & 29.75 & 30.80 & \underline{34.26} & 36.41 & 38.55 & 37.56 & 40.26 & 42.95 \\
    Qwen3.5-A3B & ours & 37.04 & 39.53 & 42.02 & 38.43 & 42.10 & 45.76 & 33.32 & 35.48 & 37.65 & 33.28 & 35.68 & 38.08 & 35.52 & 38.20 & 40.88 \\
    \hline
    \multicolumn{17}{l}{\textit{Streaming Video Models (Trained for Streaming Responses)}} \\
    \hline
    VideoLLM-Online & default & 26.20 & 26.21 & 26.22 & 25.08 & 25.10 & 25.13 & 16.92 & 17.46 & 18.00 & 25.00 & 25.00 & 25.00 & 23.30 & 23.44 & 23.59 \\
    EyeWO & default & 30.25 & 32.03 & 33.81 & 39.84 & 41.32 & 42.79 & 20.91 & 20.48 & 20.05 & 22.49 & 21.41 & 20.33 & 28.37 & 28.81 & 29.25 \\
    MMDuet & default & 43.13 & 46.97 & 50.81 & 41.51 & 43.62 & 45.73 & 22.13 & 23.14 & 24.15 & 24.18 & 24.36 & 24.53 & 32.74 & 34.52 & 36.31 \\
    MMDuet2 & default & \underline{43.26} & \underline{52.48} & \underline{61.71} & 27.98 & 28.99 & 30.00 & \underline{37.50} & \underline{44.10} & \textbf{50.70} & 28.41 & 30.56 & 32.71 & 34.29 & 39.03 & 43.78 \\
    VideoChat3 & default & 28.74 & 30.26 & 31.78 & 29.00 & 30.77 & 32.54 & 32.98 & 37.44 & 41.90 & 27.03 & 28.23 & 29.44 & 29.44 & 31.68 & 33.91 \\ \rowcolor{techgray}
    \gemmaModel (ours) & ours & 40.00 & 45.26 & 50.51 & 41.42 & 45.52 & 49.62 & 35.36 & 39.68 & 44.01 & 34.16 & \underline{37.06} & \underline{39.95} & \underline{37.74} & \underline{41.88} & \underline{46.02} \\ \rowcolor{techgray}
    \qwenModel (ours) & ours & 42.57 & 48.43 & 54.29 & \underline{41.63} & \underline{45.96} & \underline{50.29} & \textbf{40.43} & \textbf{45.47} & \underline{50.50} & \textbf{36.47} & \textbf{40.43} & \textbf{44.39} & \textbf{40.28} & \textbf{45.07} & \textbf{49.87} \\

    \bottomrule
\end{tabular}
}\end{table}

\paragraph{Extended Training Comparisons.}
\Cref{tab:proactive_videoqa_subset,tab:proactive_videoqa_method} extend the controlled comparisons in \Cref{tab:controlled_training} to $\omega=0.0$, $0.5$, and $1.0$.
For the fixed-recipe data ablation, both external datasets are converted into our \texttt{Silence}, \texttt{Response}, and \texttt{Hold} targets: each original response is kept, the next two seconds are labeled \texttt{Hold}, and the source labels then resume. For VideoChat3, its standby state is additionally mapped to \texttt{Silence}. Mean averages the four splits.

\begin{table}[t]
\caption{
\textbf{Training-data comparison on ProactiveVideoQA.}
We report PAUC-F for the pretrained Qwen3.5-A3B checkpoint (Base Model) and 3 checkpoints post-trained on datasets from MMDuet2 \citep{wang2026mmduet2}, VideoChat3 \citep{li2026videochat3}, and our proposed \dataset dataset.
Our \dataset dataset demonstrates better suitability as a post-training dataset for proactive streaming usecases. 
\textbf{Best} and \underline{second-best} marked in each column.
}
\vspace{-0.5em}
\label{tab:proactive_videoqa_subset}
\centering
\small
\def\arraystretch{1.3}  
\setlength\tabcolsep{0.7em}  
\scalebox{0.70}{%
\begin{tabular}{l|ccc|ccc|ccc|ccc|ccc}
    \toprule
    Dataset & \multicolumn{3}{c|}{\texttt{[WEB]}} & \multicolumn{3}{c|}{\texttt{[EGO]}} & \multicolumn{3}{c|}{\texttt{[TV]}} & \multicolumn{3}{c|}{\texttt{[VAD]}} & \multicolumn{3}{c}{\texttt{[Mean]}} \\
    \hline
    $\omega=$ & 0.0 & 0.5 & 1.0 & 0.0 & 0.5 & 1.0 & 0.0 & 0.5 & 1.0 & 0.0 & 0.5 & 1.0 & 0.0 & 0.5 & 1.0 \\
    \hline
    \hline
    Base Model & 51.85 & 54.24 & 56.54 & 37.62 & 39.29 & 40.82 & 33.42 & 34.47 & 35.46 & 41.90 & 43.75 & 45.51 & 41.20 & 42.94 & 44.58 \\
    MMDuet2 & 26.72 & 27.34 & 27.89 & 23.64 & 24.39 & 25.21 & 16.31 & 16.49 & 16.72 & 7.42 & 7.46 & 7.82 & 18.52 & 18.92 & 19.41 \\
    VideoChat3 & \underline{52.30} & \underline{55.15} & \underline{56.92} & \underline{38.24} & \underline{39.71} & \underline{41.52} & \underline{39.52} & \underline{40.53} & \underline{43.82} & \underline{43.24} & \underline{45.72} & \underline{47.45} & \underline{43.33} & \underline{45.28} & \underline{47.43} \\ \rowcolor{techgray}
    \dataset & \textbf{53.13} & \textbf{57.48} & \textbf{61.41} & \textbf{43.77} & \textbf{46.05} & \textbf{48.13} & \textbf{40.71} & \textbf{43.11} & \textbf{45.25} & \textbf{46.56} & \textbf{49.67} & \textbf{52.55} & \textbf{46.04} & \textbf{49.08} & \textbf{51.84} \\
    \bottomrule
\end{tabular}
}\end{table}

\begin{table}[t]
\caption{
\textbf{Training Backbone Ablation.}
We report PAUC-F for MMDuet2 and VideoChat3 with a common Qwen3.5-A3B backbone as our \qwenModel.
The backbone is held fixed, so the comparison reflects our training method and dataset.
\textbf{Best} and \underline{second-best} marked in each column.
}
\vspace{-0.5em}
\label{tab:proactive_videoqa_method}
\centering
\small
\def\arraystretch{1.3}  
\setlength\tabcolsep{0.7em}  
\scalebox{0.70}{%
\begin{tabular}{l|ccc|ccc|ccc|ccc|ccc}
    \toprule
    Method & \multicolumn{3}{c|}{\texttt{[WEB]}} & \multicolumn{3}{c|}{\texttt{[EGO]}} & \multicolumn{3}{c|}{\texttt{[TV]}} & \multicolumn{3}{c|}{\texttt{[VAD]}} & \multicolumn{3}{c}{\texttt{[Mean]}} \\
    \hline
    $\omega=$ & 0.0 & 0.5 & 1.0 & 0.0 & 0.5 & 1.0 & 0.0 & 0.5 & 1.0 & 0.0 & 0.5 & 1.0 & 0.0 & 0.5 & 1.0 \\
    \hline
    \hline
    MMDuet2 & 31.82 & 32.76 & 33.54 & 28.95 & 29.57 & 30.70 & 21.36 & 22.20 & 22.54 & 12.88 & 13.03 & 13.42 & 23.75 & 24.39 & 25.05 \\
    VideoChat3 & \underline{34.67} & \underline{35.86} & \underline{36.88} & \underline{30.12} & \underline{30.36} & \underline{31.34} & \underline{37.97} & \underline{40.75} & \underline{42.84} & \underline{24.28} & \underline{24.67} & \underline{24.88} & \underline{31.76} & \underline{32.91} & \underline{33.99} \\ \rowcolor{techgray}
    \qwenModel & \textbf{53.13} & \textbf{57.48} & \textbf{61.41} & \textbf{43.77} & \textbf{46.05} & \textbf{48.13} & \textbf{40.71} & \textbf{43.11} & \textbf{45.25} & \textbf{46.56} & \textbf{49.67} & \textbf{52.55} & \textbf{46.04} & \textbf{49.08} & \textbf{51.84} \\
    \bottomrule
\end{tabular}
}
\end{table}

\begin{table}[t]
\caption{
\textbf{ESTP-F1 score on Full ESTP-Bench.}
We report complete per-task ESTP-Bench results corresponding to the category-level summary in \Cref{tab:ESTP_result} of our main paper. Baselines are from \citet{zhang2025eyeswideopenego} and our 4 models evaluated following identical settings. 
\textbf{Best} and \underline{second-best} are marked in each column.}
\label{tab:ESTP_result_full}
\centering
\resizebox{\textwidth}{!}{%
\begin{tabular}{l| ccccccccc |ccccc |ccc |c}
\toprule
\multirow{2}{*}{\textbf{Model}}  & \multicolumn{9}{c|}{\textbf{\color{techdark}Explicit Proactive Task}} & \multicolumn{5}{c|}{\textbf{\color{techdark}Implicit Proactive Task}} & \multicolumn{3}{c|}{\textbf{\color{techdark}Contextual Q}} & \multirow{2}{*}{\textbf{\color{techdark}Overall}}  \\
\cmidrule(lr){2-10}  \cmidrule(lr){11-15}  \cmidrule(lr){16-18}
 & OR & AP & TRU & OL & OSC & EOL & EOSC & AR & All & OFR & IFR & NAR & TU & All & ORC & TRC & All & \\
\midrule[1pt] \rowcolor{techgray!50}
\multicolumn{19}{l}{\cellcolor{techgray!30}\textit{Offline MLLMs Response-in-Last}} \\
LLaVA-OneVision  & 7.2 & 11.5 & 4.9 & 10.0 & 4.9 & 6.9 & 5.6 & 3.2 & 6.8 & 3.8 & 6.3 & 11.6 & 29.8 & 12.9 & 10.8 & 5.7 & 8.2 & 8.7 \\
Qwen2-VL  & 11.7 & 8.1 & 14.9 & 10.5 & 1.7 & 8.9 & 10.6 & 6.0 & 9.0 & 10.2 & 4.4 & 26.5 & 49.5 & 22.6 & 13.3 & 9.4 & 11.3 & 13.3 \\
MiniCPM-V   & 12.3 & 12.6 & 10.7 & 13.7 & 8.6 & 7.5 & 11.9 & 5.5 & 10.4 & 11.8 & 9.2 & 36.0 & 55.3 & 28.1 & 32.6 & 25.4 & 29.0 & 18.1 \\
LLaVA-NeXT-Video  & 8.3 & 9.4 & 7.4 & 10.2 & 7.8 & 7.4 & 10.3 & 5.6 & 8.3 & 6.4 & 6.7 & 21.1 & 45.9 & 20.0 & 10.1 & 9.8 & 9.9 & 11.9 \\
InternVL-V2 & 9.3 & 14.6 & 9.5 & 10.6 & 1.7 & 6.3 & 3.0 & 3.6 & 7.3 & 3.3 & 9.2 & 15.5 & 28.2 & 14.0 & 16.9 & 15.6 & 16.2 & 10.5 \\
\midrule[0.8pt] \rowcolor{techgray!30}
\multicolumn{19}{l}{\cellcolor{techgray!30}\textit{VLMs for Streaming Detection}} \\
CLIP  & 7.3 & 9.5 & 7.4 & 8.5 & 1.8 & 4.7 & 2.2 & 2.7 & 5.5 & 2.8 & 5.2 & 51.3 & 29.3 & 22.2 & 4.6 & 3.8 & 4.2 & 10.1 \\
LaViLa   & 8.4 & 10.7 & 9.0 & 9.1 & 3.1 & 5.4 & 3.6 & 4.3 & 6.7 & 7.8 & 10.0 & 56.2 & 34.4 & 27.1 & 9.4 & 28.9 & 19.2 & 14.3 \\
EgoVLP  & 10.5 & 11.0 & 8.7 & 8.5 & 5.5 & 5.6 & 5.3 & 4.4 & 7.4 & 6.2 & 10.7 & \underline{58.4} & 48.3 & 30.9  & 8.0 & 25.3 & 16.6 & 15.5 \\
\midrule[0.8pt] \rowcolor{techgray!30}
\multicolumn{19}{l}{\cellcolor{techgray!30}\textit{Offline MLLMs Polling Strategy}} \\
LLaVA-OneVision  & 8.3 & 8.8 & 22.8 & 25.4 & 13.5 & 9.8 & 9.6 & 10.3 & 13.6 & 20.3 & 20.9 & 35.9 & 49.9 & 31.8 & 14.6 & 1.9 & 8.2 & 18.0 \\
Qwen2-VL  & 13.7 & 13.5 & 15.4 & 29.5 & 8.0 & 15.4 & 16.6 & 10.9 & 15.4 & 17.8 & 19.8 & 56.4 & 63.1 & \underline{39.3} & 13.0 & 7.7 & 10.4 & 21.3 \\
MiniCPM-V  & 14.9 & 16.8 & 17.1 & 26.8 & 7.7 & 12.9 & 12.5 & 13.1 & 15.2 & 15.9 & 21.0 & 46.8 & 62.2 & 36.5 & 24.3 & 28.9 & 26.6 & 22.9 \\
LLaVA-NeXT-Video & 15.6 & 14.6 & 21.9 & 26.8 & 12.8 & 14.2 & 13.5 & 12.3 & 16.5 & 18.6 & 23.2 & 44.9 & 51.6 & 34.6 & 19.9 & 7.7 & 13.8 & 21.3 \\
InternVL-V2  & 11.3 & 5.9 & 7.0 & 10.1 & 0.7 & 2.7 & 5.2 & 2.2 & 5.6 & 8.3 & 2.9 & 4.3 & 11.2 & 6.7 & 6.1 & 5.3 & 5.7 & 5.9 \\
\midrule[0.8pt] \rowcolor{techgray!30}
\multicolumn{19}{l}{\cellcolor{techgray!30}\textit{Online MLLMs}} \\
LIVE(threshold=0.8) & 9.7 & 11.0 & 7.4 & 10.8 & 1.9 & 6.0 & 3.6 & 5.6 & 7.0 & 4.2 & 7.4 & 12.9 & 12.8 & 9.3 & 19.6 & 13.8 & 11.8 & 9.1 \\
LIVE(threshold=0.9)  & 11.2 & 13.9 & 7.9 & 13.2 & 5.6 & 9.4 & 6.0 & 8.9 & 9.5 & 5.8 & 8.9 & 41.0 & 46.7 & 25.6 & 11.3 & 26.5 & 18.9 & 15.5 \\
MMDuet  & 7.2 & 10.3 & 17.6 & 10.2 & 4.2 & 6.1 & 8.8 & 8.5 & 9.1 & 10.0 & 7.7 & 50.1 & \underline{69.1} & 34.2 & 17.4 & 23.1 & 20.3 &  17.8 \\
EyeWO & 26.6 & \underline{26.6} & 25.1 & 26.8 & 19.8 & 22.3 & 20.8 & 20.7 & 23.6 & 24.8 & \textbf{31.0} & \textbf{75.3} & \textbf{78.7} & \textbf{52.5} & 39.5 & \textbf{47.8} & \textbf{43.6} & \underline{34.7}\\
\midrule[0.8pt] \rowcolor{techgray!30}
\multicolumn{19}{l}{\cellcolor{techgray!30}\textit{Ours}} \\
Gemma4-A4B & 19.4 & 23.5 & 24.0 & 20.9 & 10.6 & 16.8 & 13.1 & 17.4 & 18.2 & 12.6 & 16.1 & 44.1 & 52.0 & 31.2 & 21.0 & 33.3 & 27.1 & 23.2 \\
Qwen3.5-A3B & 14.4 & 15.5 & 22.9 & 19.7 & 16.9 & 12.2 & 20.0 & 11.1 & 16.6 & 19.9 & 14.4 & 14.5 & 41.8 & 22.6 & 16.1 & 12.2 & 14.1 & 18.0 \\
\gemmaModel & \underline{30.3} & 24.7 & \textbf{38.9} & \underline{33.5} & \underline{28.3} & \underline{31.5} & \underline{32.8} & \underline{32.9} & \underline{31.6} & \underline{26.4} & \underline{30.5} & 28.4 & 36.3 & 30.4 & \underline{41.4} & \underline{39.6} & \underline{40.5} & 32.5 \\
\qwenModel & \textbf{32.8} & \textbf{31.8} & \underline{36.9} & \textbf{35.6} & \textbf{32.2} & \textbf{34.5} & \textbf{38.6} & \textbf{39.1} & \textbf{35.2} & \textbf{27.6} & 30.4 & 25.3 & 45.3 & 32.1 & \textbf{45.8} & 34.9 & 40.4 & \textbf{35.1} \\
\bottomrule
\end{tabular}%
  }\end{table}

\begin{table}[t]
\centering
\begin{minipage}{\linewidth}
\caption{
\textbf{PAUC-F on Full ESTP-Bench.}
We report per-task PAUC-F ($\omega=0.5$, GPT-4.1 judge) on the full ESTP-Bench.
\textbf{Best} and \underline{second-best} are marked in each column.
EyeWO \citep{zhang2025eyeswideopenego} official checkpoint evaluated under identical settings. 
}
\label{tab:estp_paucf_full}
\vspace{-0.5em}
\centering
\def\arraystretch{1.3}
\setlength\tabcolsep{0.6em}
\scalebox{0.55}{%
\begin{tabular}{l| ccccccccc |ccccc |ccc |c}
\toprule
\multirow{2}{*}{\textbf{Model}}  & \multicolumn{9}{c|}{\textbf{\color{techdark}Explicit Proactive Task}} & \multicolumn{5}{c|}{\textbf{\color{techdark}Implicit Proactive Task}} & \multicolumn{3}{c|}{\textbf{\color{techdark}Contextual Q}} & \multirow{2}{*}{\textbf{\color{techdark}Overall}}  \\
\cmidrule(lr){2-10}  \cmidrule(lr){11-15}  \cmidrule(lr){16-18}
 & OR & AP & TRU & OL & OSC & EOL & EOSC & AR & All & OFR & IFR & NAR & TU & All & ORC & TRC & All & \\
\midrule[1pt]
EyeWO & 10.86 & 7.08 & 3.43 & 10.72 & 7.16 & 10.20 & 4.56 & 11.23 & 8.16 & 17.48 & 18.35 & 5.76 & 6.80 & 12.10 & 16.22 & 6.64 & 11.43 & 9.75 \\
Gemma4-A4B & 11.05 & 16.90 & 15.59 & 14.07 & 7.52 & 11.16 & 6.41 & 10.33 & 11.63 & 9.66 & 10.70 & 9.92 & 13.43 & 10.93 & 11.51 & 14.02 & 12.77 & 11.59 \\
Qwen3.5-A3B & \underline{23.88} & \underline{27.17} & 23.07 & \underline{29.78} & \underline{29.24} & 23.10 & \underline{29.87} & \underline{23.47} & \underline{26.20} & \textbf{29.49} & 23.28 & \textbf{31.65} & 35.15 & \underline{29.90} & \underline{35.67} & \underline{28.46} & \underline{32.07} & \underline{28.09} \\ 
\gemmaModel & 20.23 & 25.97 & \underline{30.06} & \textbf{29.83} & 20.89 & \underline{24.41} & 17.73 & 21.28 & 23.80 & 23.43 & \underline{30.27} & 26.51 & \underline{35.95} & 29.04 & 32.36 & 25.61 & 28.99 & 26.04 \\ 
\qwenModel & \textbf{24.62} & \textbf{31.77} & \textbf{30.47} & 29.08 & \textbf{30.82} & \textbf{27.84} & \textbf{31.64} & \textbf{24.24} & \textbf{28.81} & \underline{28.56} & \textbf{36.56} & \underline{27.07} & \textbf{36.42} & \textbf{32.15} & \textbf{38.87} & \textbf{29.50} & \textbf{34.19} & \textbf{30.53} \\
\bottomrule
\end{tabular}%
}

\vspace{1.5em}
\end{minipage}
\begin{minipage}{\linewidth}
\caption{
\textbf{Golden Fires per query on Full ESTP-Bench.}
We report mean assistant fires per query, so repeated and rephrased responses both count.
A query denotes one question on \textit{Explicit} and \textit{Implicit} splits, and one multi-question session on \textit{Contextual} split.
The first row calculates Gold Rate by assigning a single response to each gold window, which is how an ideal assistant is expected to respond.}
\label{tab:estp_fires}
\vspace{-0.5em}
\centering
\def\arraystretch{1.3}
\setlength\tabcolsep{0.6em}
\scalebox{0.55}{%
\begin{tabular}{l| ccccccccc |ccccc |ccc |c}
\toprule
\multirow{2}{*}{\textbf{Model}}  & \multicolumn{9}{c|}{\textbf{\color{techdark}Explicit Proactive Task}} & \multicolumn{5}{c|}{\textbf{\color{techdark}Implicit Proactive Task}} & \multicolumn{3}{c|}{\textbf{\color{techdark}Contextual Q}} & \multirow{2}{*}{\textbf{\color{techdark}Overall}}  \\
\cmidrule(lr){2-10}  \cmidrule(lr){11-15}  \cmidrule(lr){16-18}
 & OR & AP & TRU & OL & OSC & EOL & EOSC & AR & All & OFR & IFR & NAR & TU & All & ORC & TRC & All & \\
\midrule[1pt]
Gold Rate & 3.98 & 4.63 & 2.22 & 2.67 & 4.77 & 3.45 & 2.10 & 6.29 & 4.07 & 2.73 & 2.73 & 5.34 & 6.34 & 4.45 & 2.86 & 15.10 & 8.87 & 5.41 \\
\midrule
EyeWO & 15.43 & 17.08 & 8.83 & 11.23 & 25.45 & 17.36 & 11.82 & 23.73 & 17.12 & 13.81 & 10.41 & 96.02 & 109.93 & 62.61 & 19.01 & 234.29 & 124.65 & 55.28 \\
Gemma4-A4B & 19.72 & 12.96 & 14.25 & 17.79 & 33.29 & 25.57 & 33.45 & 39.48 & 25.65 & 26.15 & 19.60 & 49.41 & 32.92 & 33.33 & 23.03 & 48.90 & 35.75 & 29.98 \\
Qwen3.5-A3B & 1.02 & 0.92 & 0.96 & 1.02 & 0.86 & 1.24 & 1.10 & 1.06 & 1.04 & 1.34 & 1.21 & 0.84 & 3.81 & 1.86 & 0.58 & 2.19 & 1.37 & 1.31 \\
\gemmaModel & 4.97 & 5.01 & 2.36 & 4.51 & 7.00 & 6.67 & 5.07 & 7.05 & 5.53 & 3.68 & 3.19 & 4.21 & 4.25 & 3.90 & 4.29 & 9.72 & 6.96 & 5.55 \\
\qwenModel & 4.57 & 3.58 & 1.89 & 4.14 & 3.09 & 3.58 & 2.87 & 5.73 & 3.94 & 2.29 & 1.75 & 3.83 & 5.20 & 3.43 & 3.17 & 8.29 & 5.69 & 4.29 \\
\bottomrule
\end{tabular}%
}

\vspace{1.5em}
\end{minipage}
\begin{minipage}{\linewidth}
\caption{
\textbf{Gold-window lengths across ESTP-Bench splits.}
Explicit, Implicit, and Contextual items use source videos of similar length, but their gold windows differ sharply.
Compared to explicit, implicit gold windows are $3.85\times$ longer and contextual gold windows are $3.14\times$ longer.
}
\label{tab:estp_split_windows}
\vspace{-0.5em}
\centering
\small
\def\arraystretch{1.15}
\setlength\tabcolsep{1.2em}
\scalebox{0.85}{%
\begin{tabular}{@{}lrrrr@{}}
\toprule
Split & Queries & \begin{tabular}{@{}c@{}}Average\ \\Video Length\end{tabular} & \begin{tabular}{@{}c@{}}Average \\Gold Window\end{tabular} & \begin{tabular}{@{}c@{}}Total Gold-Window \\ Time / Query \end{tabular} \\
\midrule \rowcolor{techgray}
Explicit & 851 & 29:09 & 4.81\,s & 19.99\,s \\
Implicit & 361 & 30:29 & 18.53\,s & 84.07\,s \\
Contextual & 430 & 33:34 & 15.08\,s & 133.70\,s \\
\bottomrule
\end{tabular}%
}

\end{minipage}
\end{table}

\subsection{Harness Ablations}
\label{app:sub_ablate}

We next provide additional ablations on our harness design. 
We first compare against existing harness designs from \citet{wang2025streambridge,yao2026harnessing} in \Cref{tab:harness_comparison}. 
Then we compare against harness tuning with a general strategy built following \citet{zhang2026darwinx} without our explicit separation of harness components used in our tuning pipeline. 
This comparison is also included in \Cref{tab:harness_comparison}. 
These results demonstrate how our gains through automated harness tuning are non-trivial and result in a harness outperforming prior work on proactive streaming tasks. 

\begin{figure}[t]
  \centering
  \includegraphics[width=\linewidth]{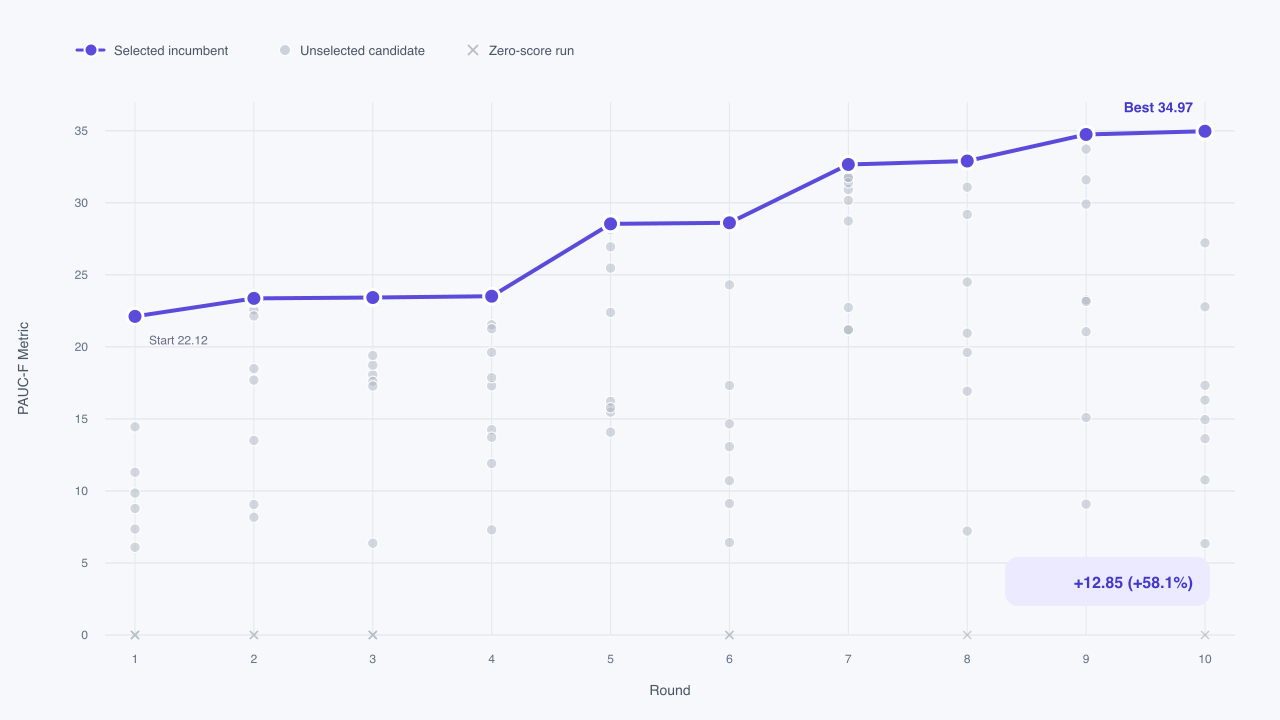}
  \caption{
  \textbf{Automated Harness Tuning.}
  We plot all 100 evaluations on our \dataset validation with various harness design candidates for Gemma4-A4B model.
  The selected models for each round are highlights (blue).
  The full tuning process achieves a gain of 12.85 PAUC-F points on the validation data. 
  }
  \label{fig:harness_search_progress}
\end{figure}

We also report the full numbers for harness tuning on the \dataset validation set for our Gemma4-A4B model in \Cref{fig:harness_search_progress}.

\begin{table}[t]
\centering
\small
\caption{
\textbf{Additional Ablation on Harness Design.}
PAUC-F ($\omega=0.5$) on WEB and EGO for Gemma4-A4B.
We compare our harness against a harness tuned by general harness evolution~\citep{zhang2026darwinx}, and harnesses from StreamBridge~\citep{wang2025streambridge} and \citet{yao2026harnessing}.
All baselines are implemented by us and evaluated under identical settings.
}
\label{tab:harness_comparison}
\vspace{-0.5em}
\def\arraystretch{1.3}
\setlength\tabcolsep{1.2em}
\scalebox{0.90}{%
\begin{tabular}{@{}lcc@{}}
\toprule
Method & WEB & EGO \\
\midrule
\rowcolor{techgray}
Ours (Default)              & 43.4 & 36.0 \\
StreamBridge Harness        & 32.5 &  6.2 \\
Streaming Harness           & 28.5 & 12.7  \\
General Harness Evolution   & 40.2 & 31.4 \\
\bottomrule
\end{tabular}
}
\end{table}

Both backbone-specific searches converge to the same high-level design despite different pretraining, and the final search iteration changes performance only slightly (\Cref{fig:harness_search_progress}). This supports convergence across these two backbones, rather than establishing stability across random validation subsets. Longer visual windows did not improve the frozen-model harness; finetuned models benefited from context up to 32 frames. We fix the decision rate at 1\,Hz to match prior work; higher rates could improve temporal resolution but were not evaluated.

\section{Interactive PO Task}
\label{app:po_interactive}

Existing proactive evaluation tasks contain a single query asked at start of the video and their established inference strategy does not involve injecting new queries mid video.
We introduce our PO-I task using the same videos in the PO split of StreamingBench \citep{lin2024streamingbench}, where each video has up to 5 queries injected during the video. 
This requires agentic models to handle multiple queries at different times of the videos, introducing several new challenges. 

We build each PO-I episode from one video in the StreamingBench PO split. The video is streamed once, and its five annotated queries are injected at their original timestamps. We score every query with the same time-accuracy rule as the official PO split.
The key difference lies in the inference setting, which injects queries interleaved with video frames, making the task more difficult. 
In these videos, each next query arrives only after the preceding gold response, so PO-I evaluates completed-request updates. The model must switch to the newly injected query while earlier frames and responses remain in context, and must keep later responses specific to that query.

We note that our training videos do not contain this style of queries mid video. 
However, our memory design allows model to condition its behavior on the current memory state. 
This is enforced in the harness and training as well. 
This allows the model to generalize zero-shot to multi-query interactions during inference, which aligns with positive results we obtain for this task.

\section{Additional Dataset Details}
\label{app:dataset_documentation}

For all experiments, \dataset train and validation splits were filtered for overlap with ProactiveVideoQA, ESTP-Bench, StreamingBench (PO), and OVOBench forward-active-responding tasks, by matching normalized source-video IDs (including parent-video IDs) and file checksums.

\paragraph{Dataset Splits.}
We report statistics on the data splits of \dataset dataset used for our training and validation in \Cref{tab:dataset_stats}.

\begin{table}[ht]
\caption{
\textbf{\dataset Additional Details.}  
Training and validation split details of our \dataset dataset.}
\label{tab:dataset_stats}
\vspace{-0.5em}
\centering
\small
\def\arraystretch{1.2}
\setlength\tabcolsep{1.2em}
\scalebox{0.9}{
\begin{tabular}{@{}lrrr@{}}
\toprule
Split & Videos & Questions & Responses \\
\midrule
Train & 176,021 & 176,021 & 777,278 \\
Validation & 1,768 & 1,768 & 7,666 \\
\midrule
Total & 177,789 & 177,789 & 784,944 \\
\bottomrule
\end{tabular}
}
\end{table}

\subsection{Dataset Generation Details}
\label{app:sub_dataset}

\paragraph{Automated Proactive Query-Response Generation:}

For a given video \(V\), we automatically generate a proactive interaction in two successive model calls. 

First, an instruction-generation model observes the complete video and synthesizes a single video-specific user query \(q\). The query requests notifications about a concrete and visually verifiable event, action, or state change (e.g., an object being picked up or a person entering the scene). We require each query to define a specific, visually observable trigger, rather than requesting general video narration or relying on subjective criteria such as “interesting” events.

Next, a response-generation model receives \(V\) and \(q\) and produces a temporally ordered sequence \(\mathcal{R}=\{(t_i,r_i)\}_{i=1}^{N}\), where \(t_i\) is the trigger time and \(r_i\) is a concise natural-language response. To approximate real-time deployment, we instruct the model to emulate an online observer. Although the complete video is provided to the model, we prompt the model and require it to emulate causal streaming observation: each response must rely only on evidence visible up to \(t_i\), must not anticipate future events, and must be emitted at the first frame at which the query condition is satisfied. This is a prompt-level constraint rather than a restriction on visual access; human reviewers correct responses that precede sufficient visual evidence. The model remains silent between triggers and returns an empty sequence when no valid trigger is detected throughout the video. 

To balance computational cost with the need for fine-grained temporal reasoning, we use different models for the two generation steps. Specifically, we use \texttt{gemini-3-flash-preview} for query generation and \texttt{gemini-3-pro-preview} for response generation. We remove the audio track and sample each video at 1 FPS. Consequently, trigger times are annotated at one-second resolution. The generated queries, responses, timestamps, and associated metadata are stored in structured JSON format and subsequently verified and refined during the human-review stage.

\paragraph{Human Annotation Repair:}

Human annotators first assess the validity of each generated query. A valid query must specify a concrete and visually recognizable condition that can be detected causally from a streaming video and can therefore support a well-defined response decision. We reject queries whose triggers are ambiguous, subjective, not visually observable, or dependent on future context. Only samples with valid queries proceed to response-level annotation repair.

For each valid query, annotators inspect the timing and content of every generated response and repair the annotations when necessary. A valid response must be relevant to the query, visually grounded at its assigned timestamp, free of hallucinated content, and consistent with causal observation. In particular, the response should be unambiguously supported by the frames observed up to that time, while it may use preceding context but must not reveal information from future frames. Annotators correct inaccurate text, shift temporally misaligned responses, remove spurious responses, and add missed responses as needed. 

For instantaneous events, the response is generated at the moment when the event becomes visually identifiable. For continuous events that persist over a period of time, annotators additionally enforce semantic continuity across consecutive responses. For example, later responses use expressions such as ``continues'' or ``still'' to indicate that the previously observed event remains ongoing.

\begin{figure}[t]
  \centering
  \includegraphics[width=0.72\linewidth]{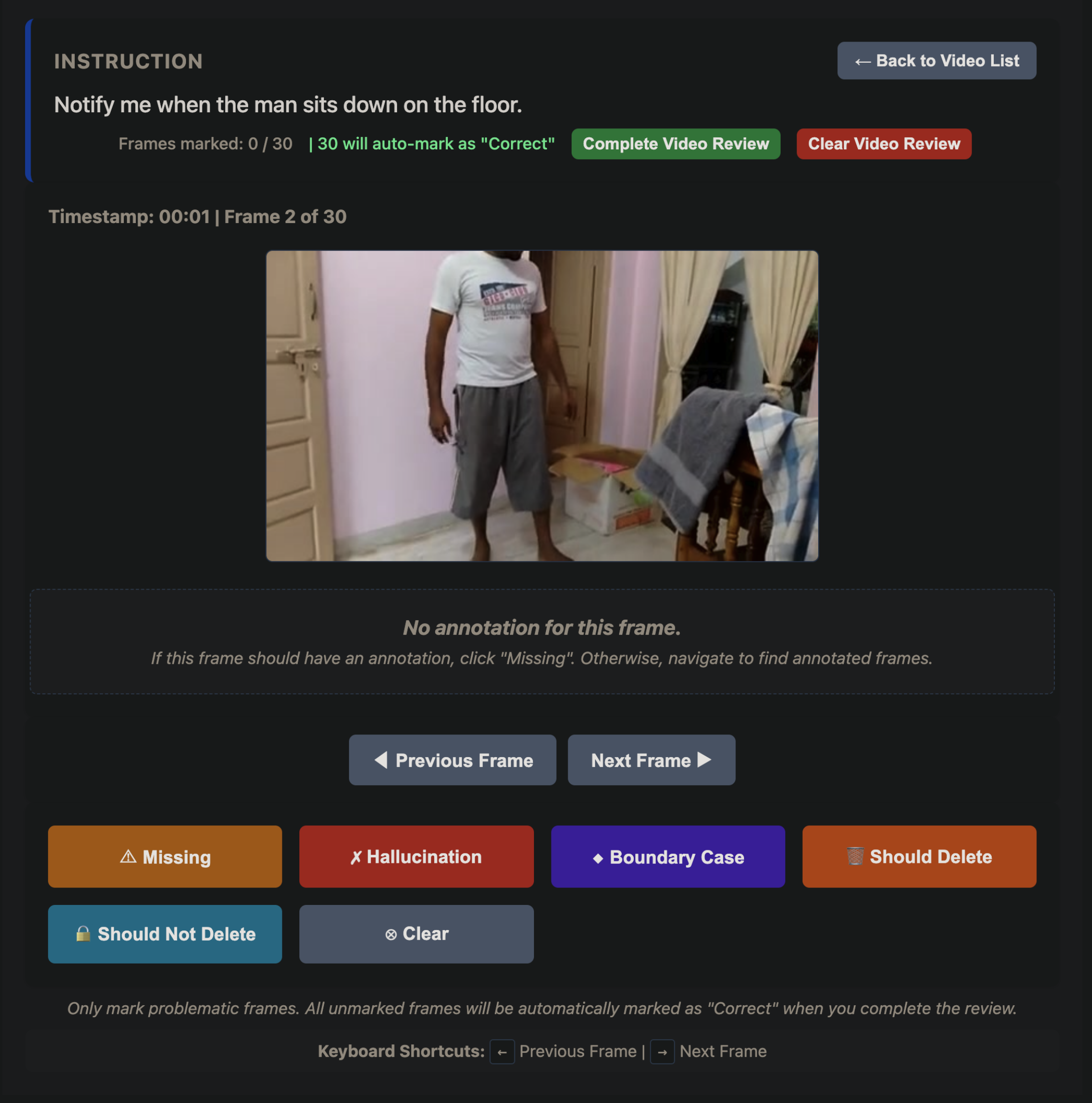}
  \caption{
  \textbf{Dataset review interface.}
  Reviewers inspect each sampled video frame against the instruction and existing annotation, navigate temporally, and mark only frames requiring correction. The interface records missing responses, hallucinations, temporal boundary cases, and deletion decisions while automatically treating unmarked frames as correct.
  }
  \label{fig:dataset_review_tool}
\end{figure}

\begin{figure}[t]
  \centering
  \includegraphics[width=\linewidth]{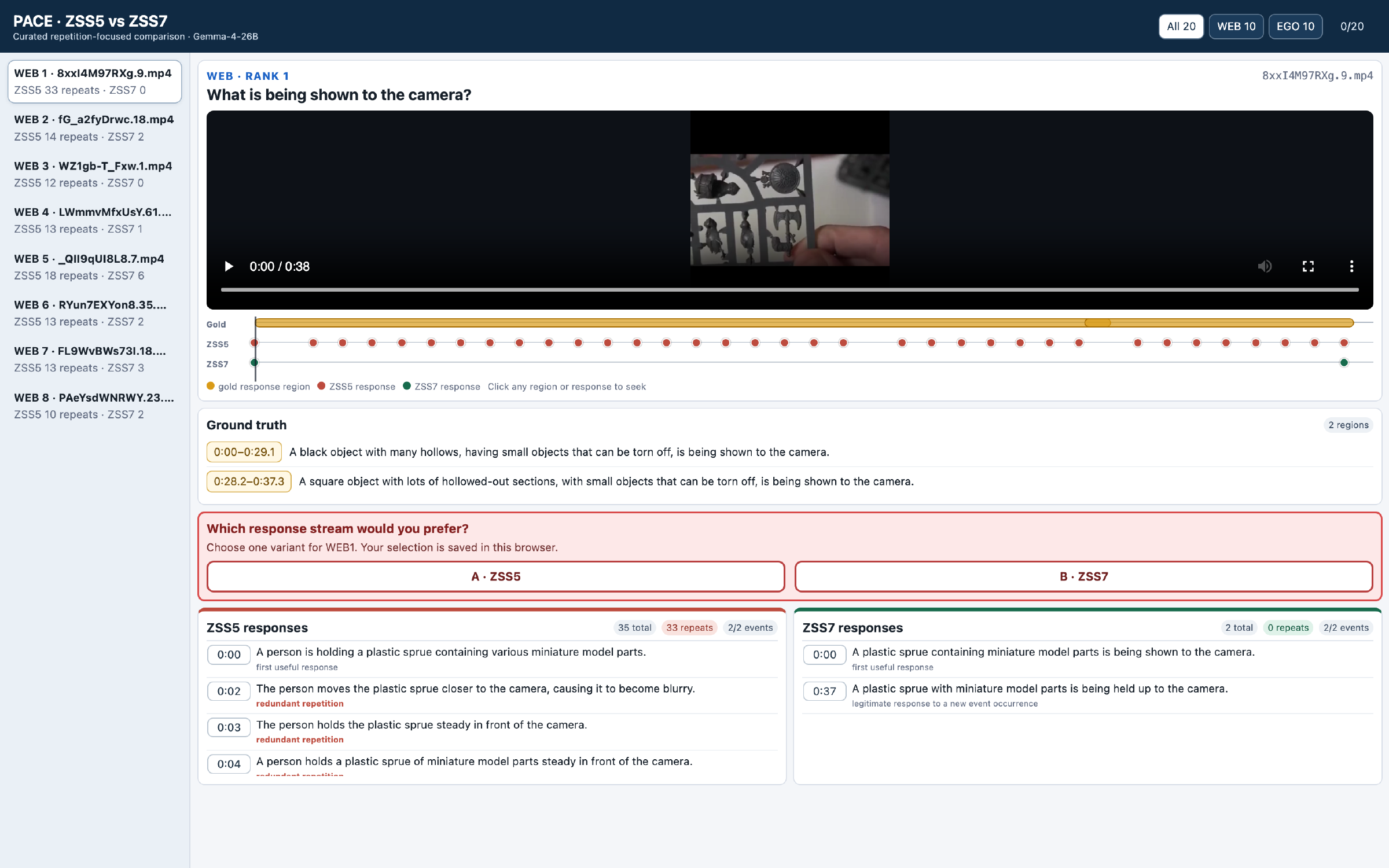}
  \caption{
  \textbf{Human rating interface for PAUC-F.}
  Raters inspect the video, ground-truth response regions, and timestamped responses from two anonymized system variants, then select the preferred response stream. The aligned timeline and side-by-side response lists expose differences in event coverage, response timing, and repetitive firing behavior.
  }
  \label{fig:paucf_human_rating_interface}
\end{figure}

\paragraph{Quality Control:}
A professional video data-annotation team recruits and trains the annotators, who receive detailed instructions and examples before performing annotation repair. We then conduct quality control at the batch level using PhD-level expert reviewers who are independent of the annotation repair process.

At the beginning of the human annotation process, we partition the automatically generated samples into ten disjoint batches and process them sequentially for human annotation repair. After the annotators complete a batch, independent expert reviewers audit a randomly selected subset of videos and their annotations using a purpose-built web interface. The reviewers examine the complete video timeline. They evaluate the annotation at each one-second time step, treating both the presence and absence of a response as a decision to assess. A response is labeled a true positive if the reviewer confirms that it is warranted and its content is correct, whereas a no-response decision is labeled a true negative if the reviewer agrees that no response is warranted. An unwarranted or incorrect response is labeled a false positive, and a missing response is labeled a false negative.

We quantify annotation quality using the F1 metric. A batch is accepted, and annotation proceeds to the next batch, only if \(F1 > 0.95\). 
Batches that do not meet this threshold undergo further correction and review. During the initial batches, we have repeated the cycle of annotator training, annotation, and independent evaluation multiple times to calibrate the annotation criteria and establish consistent quality before processing the remaining batches.

\paragraph{Dataset Review Tool}

We use the purpose-built interface shown in \Cref{fig:dataset_review_tool} for frame-level review of the generated annotations. Reviewers step through the video at one-second intervals and flag problematic frames as missing, hallucinated, boundary cases, or incorrect deletion decisions. Unmarked frames are recorded as correct when the reviewer completes the video, enabling efficient and consistent auditing of both responses and silence decisions.

\section{PAUC-F Metric Details}
\label{app:paucf}

We discuss the behavior of proposed PAUC-F compared to the parent PAUC metric and then present details on our human study. 

\paragraph{Behavior Compared to PAUC.}
PAUC \citep{wang2025proactivevqa} scores each gold turn on its reply window $(t^{\mathrm{start}}, t^{\mathrm{end}})$.
In-window replies at $\tau_1<\cdots<\tau_P$ get judge scores $s_p$, and PAUC is the area under that time--score polyline divided by the maximum area $(t^{\mathrm{end}}-t^{\mathrm{start}})S$, with $S=2$:
\begin{equation}
\label{eq:pauc}
\mathrm{PAUC}
=
\frac{
(\tau_1-t^{\mathrm{start}})\cdot 0.5
+ \sum_{p=1}^{P-1}(\tau_{p+1}-\tau_p)\, s_p
+ (t^{\mathrm{end}}-\tau_P)\, s_P
}{(t^{\mathrm{end}}-t^{\mathrm{start}})\, S}.
\end{equation}
With no in-window reply this ratio is fixed at $0.5/S=0.25$. This is the no-reply score, not a global minimum: wrong in-window replies can lower PAUC toward zero.
Before the area is taken, timestamps are shifted by $\omega\in[0,1]$,
\begin{equation}
\label{eq:pauc_omega}
\tau_p' = t^{\mathrm{start}} + (1-\omega)(\tau_p - t^{\mathrm{start}}),
\end{equation}
and $\tau_p'$ replaces $\tau_p$ in \Cref{eq:pauc}.
$\omega=0$ leaves time unchanged, so a late reply adds little area.
$\omega=1$ moves every $\tau_p$ to $t^{\mathrm{start}}$, so PAUC reduces to $s_P/S$ and timing drops out.
Our PAUC-F metric inherits the $\omega$ term from PAUC, which allows adjusting the weight between quality and timeliness.

Repeating the definition from \Cref{sec:paucf},
\begin{equation}
\label{eq:paucf}
R = \mathrm{PAUC}, \qquad
P = \frac{N_{\mathrm{value}}}{N_{\mathrm{fires}}}, \qquad
\mathrm{PAUC\text{-}F} = \frac{2PR}{P+R}.
\end{equation}
A model that never fires has $P=0$.
PAUC still returns its no-reply score $0.25$, but the harmonic mean is zero, so PAUC-F $=0$.

The judge scores responses on $\{0,1,2\}$, with maximum $S=2$ as in PAUC. Both scores 1 and 2 count as value-adding for precision, while PAUC retains their quality difference. Within each benchmark split or task, we pool firing counts across videos to compute $P$ and follow official PAUC aggregation by averaging over gold response turns to compute $R$, then take their harmonic mean. Reported cross-split Means average the split scores. We set $P=0$ when there are no fires and PAUC-F $=0$ when $P+R=0$.

Rescaling PAUC by $(\mathrm{PAUC}-0.25)/0.75$ can yield negative inputs to the harmonic mean because incorrect in-window replies can score below 0.25 (\Cref{tab:proactive_videoqa_pauc}). The sparse-policy example in \Cref{tab:sparse_policy} shows how additional fires reduce PAUC-F while leaving PAUC unchanged.

\paragraph{Human Study.}

This two-harness study tests whether excessive-firing traits are undesirable to humans; it does not establish metric rank agreement across a broad set of model pairs.

We conduct a human preference study with ten raters using the interface shown in \Cref{fig:paucf_human_rating_interface}. 
Humans rate response streams from two distinct models across 20 videos from WEB and EGO splits of ProactiveVideoQA dataset. 
Raters are instructed to select the response that seems more helpful from a real-time video assistant and are shown the video, query, and ground-truth responses (content and intervals). 

Human raters overwhelmingly select responses generated from agentic model \texttt{v2}. 
This model uses a harness tuned to maximize PAUC-F metric. 
The other \texttt{v1} model whose responses are less preferred by humans uses a harness tuned to maximize PAUC metric, often resulting in excessive repetition within windows and firing outside windows.  
The exact ratings from each human participant are provided in  \Cref{tab:human_preference_counts}.

\begin{table}[t]
\centering
\small
\caption{
\textbf{Human Preference Study.}
We present the exact results from each of the 10 human participants involved in our study. 
A clear preference for the PAUC-F optimized \texttt{v2} model was observed.
}
\label{tab:human_preference_counts}
\vspace{-0.5em}
\def\arraystretch{1.2}
\setlength{\tabcolsep}{0.7em}
\begin{tabular}{@{}ll*{10}{c}cc@{}}
\toprule
& & \multicolumn{10}{c}{Rater} & & \\
\cmidrule(lr){3-12}
Split & Choice & 1 & 2 & 3 & 4 & 5 & 6 & 7 & 8 & 9 & 10 & Total & \% \\
\midrule
\multirow{2}{*}{WEB}
 & v1 & 0 & 0 & 1 & 0 & 0 & 4 & 3 & 0 & 2 & 2 & 12 & 6.0 \\
 & v2 & 20 & 20 & 19 & 20 & 16 & 17 & 18 & 20 & 18 & 18 & 188 & 94.0 \\
\midrule
\multirow{2}{*}{EGO}
 & v1 & 0 & 1 & 0 & 0 & 2 & 2 & 0 & 0 & 1 & 0 & 6 & 3.0 \\
 & v2 & 20 & 19 & 20 & 20 & 18 & 18 & 20 & 20 & 19 & 20 & 194 & 97.0 \\
\bottomrule
\end{tabular}
\end{table}

\begin{table}[t]
  \centering
  \footnotesize
  \setlength{\abovecaptionskip}{2pt}
  \setlength{\belowcaptionskip}{0pt}
  \setlength{\aboverulesep}{0.35ex}
  \setlength{\belowrulesep}{0.35ex}
  \setlength{\tabcolsep}{3pt}
  \def\arraystretch{1.2}
  \caption{Training-free harness shared by Gemma4-A4B and Qwen3.5-A3B. Model weights stay frozen.}
  \label{tab:final_harness_config}
  \begin{tabular}{@{}p{0.20\linewidth}p{0.76\linewidth}@{}}
  \toprule
  Component & Configuration \\
  \midrule
  Decision rate & 1\,Hz \\
  Frames & Three most recent, $[t-2,t]$, truncated at the video start \\
  Timestamps & Absolute \texttt{MM:SS} immediately before each frame \\
  Question & Standing query; a later query enters only at its reveal time \\
  Memory & Last accepted response and its emission time. Acceptance replaces it; silence leaves it unchanged \\
  Decode & Greedy, at most 192 new tokens \\
  Output & \texttt{[]} for silence, or one \texttt{\{time, content\}} object \\
  Parser & Strip optional Markdown fences. Decode a JSON array, otherwise the first array-shaped span. Keep objects with a numeric \texttt{time} and non-empty \texttt{content}. Drop times outside $\pm 0.5$\,s, snap the rest to the current step, and keep at most one. Anything else is silence \\
  Gemma input & Native system, query, and observation turns. Timestamps sit beside image embeddings. Generation starts after an empty model turn \\
  Qwen input & Official chat template with thinking disabled. Query, memory, and timestamp--image pairs share one user message \\
  \bottomrule
  \end{tabular}
  
  \vspace{1.5em}
  \caption{Search and calibration budgets. A model-example evaluation is one candidate on one video.}
  \label{tab:harness_search_budget}
  \def\arraystretch{1.2}
  \begin{tabular}{@{}p{0.20\linewidth}p{0.38\linewidth}p{0.38\linewidth}@{}}
  \toprule
  Component & Full harness search & Limited calibration \\
  \midrule
  Models & Gemma and Qwen & Other offline VLMs \\
  Objective & Maximize validation PAUC-F & Maximize validation PAUC-F \\
  Editable surface & Window, memory, prompt, and output protocol & Prompt and output schema only \\
  Proposal schedule & 10 candidates per round & 1 candidate per round \\
  Budget & 10 rounds; 100 candidate evaluations/model & 5 candidates $\times$ 20 examples/model \\
  Samples per round & 500 & 20 \\
  Feedback & PAUC-F Metric & PAUC-F Metric \\
  \bottomrule
  \end{tabular}
  
  \end{table}

\begin{table}[t]
  \centering
  \footnotesize
  \setlength{\tabcolsep}{3pt}
  \caption{Training-free harness prompts for search, calibration, and inference. The frozen VLM uses a binary silence/response interface.}
  \label{tab:harness_prompts}
  \begin{tabular}{@{}p{0.20\linewidth}p{0.76\linewidth}@{}}
  \toprule
  Task & Prompt Description \\
  \midrule
  Harness tuning
  & Tune a frozen VLM harness so validation PAUC-F rises. Each round provides the current harness, one hypothesis, the editable surface, and prior scores and diagnostics. Propose the budgeted edits on that surface, and state the expected effect of each. Edits may only change the window, memory, prompt, or output protocol. Keep an edit only when validation PAUC-F improves. \\
  \addlinespace
  Harness calibration
  & Calibrate the prompt and output schema of a frozen harness so validation PAUC-F rises. Each round provides the fixed harness, the previous candidate, and its score and diagnostics. Propose one edit, and state its expected effect. Only the prompt and output schema may change. Keep an edit only when validation PAUC-F improves. \\
  \addlinespace
  Inference (Final)
  & Streaming video assistant. Each second, the input is the last accepted answer and its time (empty if none) plus the latest frames, each marked with absolute \texttt{MM:SS}. The newest stamp is now. Speak only when those frames show a useful change; otherwise return an empty list and leave the last answer as it stands. A spoken turn is one short factual sentence tagged with the current time in seconds. \\
  \bottomrule
  \end{tabular}
\end{table}

\begin{table}[t]
  \caption{Episode construction and token supervision. A non-response run is a maximal stretch of \texttt{Silence} and \texttt{Hold}.}
  \label{tab:episode_construction}
  \vspace{-0.5em}
  \centering
  \small
  \def\arraystretch{1.2}
  \begin{tabular}{@{}p{0.24\linewidth}p{0.70\linewidth}@{}}
  \toprule
  Step & Rule \\
  \midrule
  Timeline & 1\,FPS frames with absolute \texttt{MM:SS} timestamps. Consecutive frames that share a caption are one event \\
  States & First frame that supports the caption is \texttt{Response}; later frames of that event are \texttt{Hold}; every other frame is \texttt{Silence} \\
  Chunks & At most 32 frames. A cut that would split a response from its hold is moved so they stay together. Original indices and timestamps are kept \\
  Context & System instruction, standing query, and the last response that precedes the visual window \\
  Extra chunks & Response-bearing chunks are kept, together with trailing and post-response chunks so sustained silence is seen \\
  Response tokens & Every \texttt{Response} position, including its answer text \\
  Non-response run & Its first position, plus up to four further \texttt{Silence} positions and two further \texttt{Hold} positions \\
  No-response chunk & Two \texttt{Silence} positions \\
  \bottomrule
  \end{tabular}
  \end{table}

  \begin{table}[t]
  \caption{Shared optimization configuration. Adapters sit in the language tower; the visual encoder stays frozen.}
  \label{tab:training_hyperparameters}
  \centering
  \small
  \def\arraystretch{1.2}
  \begin{tabular}{@{}p{0.32\linewidth}p{0.62\linewidth}@{}}
  \toprule
  Setting & Value \\
  \midrule
  Optimizer & AdamW \\
  Learning rate & $2\times10^{-4}$ \\
  Schedule & Cosine decay; 5\% warmup \\
  Training duration & One epoch \\
  Per-device batch size & 1 \\
  Gradient accumulation & 1 \\
  Effective global batch size & 8 \\
  Precision & bfloat16 with TF32 enabled \\
  Attention & SDPA \\
  Memory optimization & Non-reentrant gradient checkpointing; ZeRO-2 \\
  Hardware & $8\times$ NVIDIA H200, one data-parallel process per GPU \\
  Random and data seeds & 42 \\
  Frames & $384\times 384$, bicubic, then the native image processor \\
  LoRA & Rank 64, scale 128, dropout 0.05 \\
  Gemma modules & Attention, dense MLP, and router \\
  Qwen modules & Attention, shared-expert MLP, and linear-recurrence projections \\
  \bottomrule
  \end{tabular}
  \end{table}

  \begin{table}[t]
    \centering
    \footnotesize
    \caption{System prompt shared by supervised finetuning and finetuned inference. The three-state output interface is used by \gemmaModel and \qwenModel.}
    \label{tab:finetuned_prompts}
    \setlength{\tabcolsep}{3pt}
    \begin{tabular}{@{}p{0.20\linewidth}p{0.76\linewidth}@{}}
    \toprule
    Task & Prompt Description \\
    \midrule
    Training and inference
    & Streaming assistant over frames labeled with absolute \texttt{MM:SS} times. Emit \texttt{</Silence>} when the frame is irrelevant, \texttt{</Hold>} when the event continues but no new answer is due, and \texttt{</Response>} plus one complete answer when the frames suffice. A prior answer may be attached; repeat it only when the new frames add evidence. Stay within what the frames show. \\
    \bottomrule
    \end{tabular}
  \end{table}

\paragraph{Theoretical Motivation.}

Our PAUC-F metric is simply the harmonic mean between two established components, the PAUC metric from \citet{wang2025proactivevqa} and firing precision. We justify our use of the harmonic mean with two key reasons.

First, the harmonic mean is the strictly correct operator for calculating the true average of rates or ratios \citep{yule1947introduction,Ferger1931TheNA}. 
In our case, both PAUC and Firing Precision represent rates of success. 
PAUC is the rate of accumulated quality-score over a bounded temporal window, while Firing Precision is the rate of valid fires over total fires. 

Second, from an optimization perspective, the harmonic mean severely penalizes high variance between its inputs \citep{van1979information}.
For streaming video agents, timeliness (PAUC) and firing restraint (Precision) are competing objectives; a model can easily maximize PAUC by spamming responses, driving Precision near zero. 
An arithmetic or geometric mean allows linear or proportional substitutability, allowing the model to \textit{hide} terrible precision behind a perfect PAUC. 
The harmonic mean enforces that the marginal utility of maximizing PAUC drops to zero if Precision is sacrificed, mathematically guaranteeing that a high PAUC-F strictly requires Pareto-optimal balance between the two.
This also mirrors the theoretical foundation of the F1 score in information retrieval.

\section{Inference Harness Details}
\label{app:harness}

This section describes the training-free harness for frozen offline VLMs.
\Cref{tab:final_harness_config} is the harness used for every reported offline model.
\Cref{tab:harness_search_budget} is the validation-only search that selected it, and
\Cref{tab:harness_prompts} presents only the harness-search, calibration, and training-free inference prompts. The frozen VLM returns silence or a response using the structured output format; it does not predict a Hold token. Prompts and state tokens for the finetuned models are described separately in \Cref{app:training}.
In proprietary-model evaluation, malformed outputs trigger up to three retries (four attempts in total) before applying the parser's silence fallback. No unresolved parsing failures remained after retries across all proprietary models tested.

During harness tuning, each round tests ten edits of one hypothesis. An edit is promoted when validation PAUC-F improves on its parent. Combined promotions are kept only after that combination itself improves, and those confirmation runs count toward the budget in \Cref{tab:harness_search_budget}.

\section{Training Pipeline Details}
\label{app:training}

\Cref{tab:finetuned_prompts} presents the shared system instruction used for supervised finetuning and inference with \gemmaModel and \qwenModel. These models predict \texttt{</Silence>}, \texttt{</Response>}, or \texttt{</Hold>}. Only \texttt{</Response>} produces a user-facing answer and updates response memory; Silence and Hold produce no answer and leave memory unchanged. Hold distinguishes an ongoing, already-reported event from irrelevant frames labeled Silence. Both training and inference use up to 32 frames and the last response as memory, at a 1\,Hz decision rate.
\Cref{tab:episode_construction} describes how we build the target episodes for supervision.
The loss is next-token prediction on the sampled state tokens and full answer text for response text tokens, all at equal weight. 
System, query, memory, visual, and unsampled target tokens stay in context only.

Examples of similar length are grouped into batches. Intermediate exposure checkpoints are saved, and the reported checkpoint is the one selected by validation PAUC-F (\Cref{sec:experiments}).

\end{document}